\documentclass[preprint,3p]{elsarticle}

\usepackage{amssymb}
\usepackage{amsmath}
\usepackage{amsfonts}
\usepackage{graphicx}
\usepackage{multirow}
\usepackage{booktabs}
\usepackage{longtable}
\usepackage{array}
\usepackage{tabularx}
\usepackage{calc}
\usepackage{textcomp}
\usepackage{xcolor}
\usepackage{url}
\usepackage{hyperref}

\journal{Applied Energy}

\newcommand{\cm}[1]{\textcolor{black}{{#1}}}

\begin{document}

\begin{frontmatter}


\title{Gated Multimodal Learning for Interpretable Property Energy Performance Prediction and Retrofit Scenario Analysis}


\author[inst1]{Yunfei Bai}
\author[inst1]{Aaron Tesfa Tsion}
\author[inst1]{Ra\'ul Rosales}
\author[inst1]{Barbara Shollock}
\author[inst1]{Wei He\corref{cor1}}
\ead{wei.4.he@kcl.ac.uk}

\cortext[cor1]{Corresponding author}

\affiliation[inst1]{organization={Department of Engineering},
            orgname={King's College London},
            city={London},
            country={UK}}

\begin{abstract}
Achieving resilient and sustainable cities requires scalable approaches to decarbonising the residential building stock, which contributes approximately 20\% of UK greenhouse gas emissions and about 25\% of energy-related emissions in the European Union. Energy Performance Certificates (EPCs) underpin regulatory and retrofit strategies, yet their reliance on on-site inspections limits city-scale assessment and timely policy response. Here we introduce a gated multimodal model that predicts continuous Standard Assessment Procedure (SAP) energy efficiency and Environmental Impact (EI) scores by fusing EPC tabular fields, assessor-written free text, and Geographic Information System-derived spatial features capturing footprint geometry, height, area, and orientation. Sample-wise gating learns property-specific modality weights, and an auxiliary band classification head stabilises training. In a case study of Westminster, London, the model predicts SAP and EI with MAE of 4.03 and 4.76 points and $R^2$ of 0.757 and 0.748, respectively (mean MAE 4.39). Modality ablation shows that full multimodal fusion improves both continuous scores and band-level accuracy over unimodal and bimodal baselines. Interpretability links predictions to decision-relevant evidence: gating weights indicate strong reliance on assessor text; SHAP highlights main fuel, built form, and construction age band; text occlusion prioritises roof and wall fields; and spatial attribution is dominated by height and footprint area, with complementary sensitivity to footprint shape. Beyond prediction, the validated framework is applied to scenario-based retrofit analysis for wall insulation, roof insulation, and window glazing upgrades, showing positive projected improvements in SAP, EI, annual energy cost, and equivalent CO$_2$ emissions. Overall, the framework provides scalable, property-level evidence for retrofit screening, intervention priorisation, and net-zero housing transitions.
\end{abstract}


\end{frontmatter}



\section{Introduction}

Improving the energy performance of existing buildings is central to climate change, energy affordability, and fuel-poverty reduction \cite{chen2023impacts,zhong2021global}. Residential buildings account for a substantial share of final energy use and carbon emissions, and many national and local decarbonisation pathways depend on the rapid identification of dwellings that should be prioritised for retrofit \cite{bai2025occupant,qu2020novel,amasyali2018review,beccali2017artificial}. In the UK, this challenge is especially acute because a large proportion of the existing housing stock will need fabric or system upgrades to meet future minimum energy standards \cite{CHEN2025100230}. Effective retrofit planning therefore requires more than generic policy targets: it requires scalable, auditable, and property-level evidence on current energy performance, likely carbon impact, and the expected benefits of candidate interventions \cite{few2023over}. 

Energy Performance Certificates (EPCs) are the main policy-facing evidence base for residential energy assessment \cite{ali2024urban}. In the UK, EPC  ratings are underpinned by the Standard Assessment Procedure (SAP), which uses building fabric, geometry, heating systems, ventilation, lighting, and fuel information to estimate energy efficiency \cite{BRE_SAP102}. EPC records also include an Environmental Impact (EI) score, which reflects the carbon implications of the dwelling \cite{BRE_SAP102}. Because EPCs are used in property markets, regulatory compliance, subsidy allocation, and retrofit planning, they are a natural foundation for urban-scale energy-performance analytics \cite{ali2020data,wang2018multi}. However, EPC assessments depend on qualified assessor inspections and are updated irregularly \cite{GOVUK_EPC_Guide}. This creates temporal gaps, incomplete coverage, and limited capacity for rapid scenario analysis across local authority housing stocks. These limitations have motivated growing interest in data-driven methods for predicting building energy performance \cite{amasyali2018review}.

\cm{Early data-driven studies show that machine learning can approximate building energy performance without repeatedly running detailed simulations. Chari and Christodoulou \cite{chari2017building}, for example, used artificial neural networks to predict Building Energy Rating classes from different levels of input detail, while Liu et al. \cite{liu2021enhancing}, Momeni et al. \cite{momeni2024enhancing}, and Olu-Ajayi et al. \cite{olu2022building} demonstrated the usefulness of random forests, neural networks, and deep learning for energy consumption prediction from envelope, thermal, and design variables. These studies establish the value of machine learning as a fast surrogate for energy simulation or assessment. However, most of them rely on single-source numerical inputs, simulated data, or building-design variables, and they generally predict energy consumption or rating classes rather than continuous policy indicators that can capture marginal retrofit gains.}

\cm{Urban-scale retrofit studies extend this line of work by linking building-stock data to decision support. Ali et al. \cite{ali2020data} developed a data-driven approach for urban retrofit decision-making in Dublin, reducing a large building-stock database to key features and recommending retrofit measures associated with target ratings. Ali et al. \cite{ali2024urban} further combined archetype development, parametric simulation, end-use demand segregation, and ensemble learning to predict urban residential building energy performance and analyse retrofit scenarios. Sheng et al. \cite{sheng2025city} presented a rapid city-scale residential energy assessment tool for Sheffield, linking spatial, morphological, and thermal characteristics to retrofit prioritisation. These studies are important because they connect prediction to retrofit planning. Yet their pipelines depend heavily on archetypes, synthetic simulation datasets, end-use segregation, or predefined building-stock variables, and the resulting models do not directly exploit the full textual and spatial evidence already contained in or linked to EPC records. Their outputs are also typically framed around energy consumption, energy use intensity, or rating transitions, rather than joint continuous SAP and EI estimation for policy-facing carbon and efficiency analysis.}

\cm{More recent work has introduced multi-source and multimodal learning into building energy assessment. Sun et al. \cite{sun2022understanding} combined EPC attributes, urban morphology, and Google Street View facade images for building energy efficiency prediction in Glasgow, improving accuracy from 79.7\% to 86.8\% when facade imagery was included. Sheng et al. \cite{sheng2022deep} used street-view images alongside EPC-derived tabular information for residential energy prediction, while Sheng et al. \cite{sheng2025learning} added transfer learning to improve prediction in data-scarce cities. These studies demonstrate that visual and administrative data can provide complementary information about building energy performance. Nevertheless, they mainly use external imagery as the additional modality, which can be affected by occlusion, image update frequency, facade visibility, privacy constraints, and limited coverage of rear or internal building conditions. They also focus primarily on energy consumption or efficiency ratings, and their fusion strategies are not designed to provide property-specific, auditable evidence about how tabular, textual, and spatial signals contribute to each prediction.}

\cm{Other recent studies have advanced interpretability and multimodal modelling in adjacent building-performance tasks. Kangalli Uyar et al. \cite{uyar2025interpretable} used XGBoost Quantile Regression and SHAP to examine how performance drivers vary across efficiency quantiles. Shen and Pan \cite{shen2023bim} integrated BIM-based simulation, explainable machine learning, and multi-objective optimisation for green building design. Li et al. \cite{li2023multimodal} proposed a multimodal GAN for matrix-based daylight prediction, Lu et al. \cite{lu2026multi} developed a cross-modal multi-task attention network using floorplans, text, numerical attributes, and topology to predict daylight and thermal comfort, and Moveh et al. \cite{moveh2025multi} combined temporal graph neural networks with weather data for multi-building operational energy forecasting. These studies show the value of explainability, uncertainty-aware modelling, spatial-temporal learning, and advanced cross-modal fusion. However, they are mainly oriented toward simulation-based design support, daylight or multi-indicator building-performance prediction, operational energy forecasting, or general performance-ratio analysis. They do not address the specific policy problem of updating and interpreting EPC-derived SAP and EI scores from routinely available property records, assessor text, and GIS-based spatial information.}

\cm{Taken together, the literature reveals four gaps. First, many studies predict energy consumption, energy use intensity, or rating bands, but fewer model continuous EPC-relevant scores that preserve marginal changes important for retrofit appraisal. Second, Environmental Impact is rarely treated as a first-class prediction target, even though carbon outcomes are central to decarbonisation policy. Third, existing multimodal approaches often rely on facade imagery or simulated design data, while the multi-field textual descriptions embedded in EPCs and the spatial context available through GIS are underused. Fourth, most fusion mechanisms provide limited sample-level interpretability, making it difficult for local authorities to understand whether a prediction is driven by tabular building attributes, assessor descriptions, or spatial/geometric context for a particular property.}

\cm{To address these gaps, this paper proposes a gated multimodal learning framework for interpretable residential energy-performance prediction and retrofit scenario analysis. The framework jointly estimates continuous SAP and EI scores using three complementary modalities: structured EPC attributes, multi-field EPC textual descriptions, and GIS-based spatial information describing property geometry and local context. A dual-target regression objective is combined with an auxiliary band-classification task so that the model learns both continuous score variation and the regulatory structure embedded in EPC bands. A sample-wise gated fusion mechanism adaptively weights the three modalities for each property, enabling the model to use different evidence combinations across heterogeneous housing stock while producing modality-level interpretability.}

\cm{This study makes three contributions. First, it develops an interpretable multimodal EPC prediction framework that jointly models tabular, textual, and spatial evidence for continuous SAP and EI estimation, moving beyond single-source predictors, image-dependent multimodal models, and band-only classification. Second, it provides a systematic evaluation in Westminster, London, including predictive performance, modality ablation, subgroup robustness, convergence behaviour, and multi-level interpretation through fusion weights, tabular feature attribution, textual field importance, and spatial contribution analysis. Third, it extends the trained model to scenario-based retrofit assessment by estimating changes in SAP, EI, annual energy cost, and equivalent carbon emissions under wall-insulation, roof-insulation, and window-glazing upgrade scenarios, thereby linking prediction accuracy to actionable local retrofit prioritisation.}

The remainder of this paper is organised as follows. Section 2 introduces the study area and the multimodal data used in this work, including data sources, pre-processing procedures, and modality construction. Section 3 presents the proposed methodology, including modality-specific encoders, the gated fusion mechanism, the prediction objective, and the construction of retrofit intervention scenarios. Section 4 evaluates the framework in terms of dataset splitting, training setup, overall predictive performance, ablation results, subgroup robustness, and convergence behaviour. Section 5 investigates model interpretability through modality-level and feature-level attribution analyses. Section 6 demonstrates the practical application of the framework through scenario-based retrofit analysis in Westminster. Finally, Section 7 summarises the main findings, discusses practical implications, and outlines the limitations and directions for future work.

\section{Study Area and Data}
\begin{figure}[htbp]
\centering
\includegraphics[width=\textwidth]{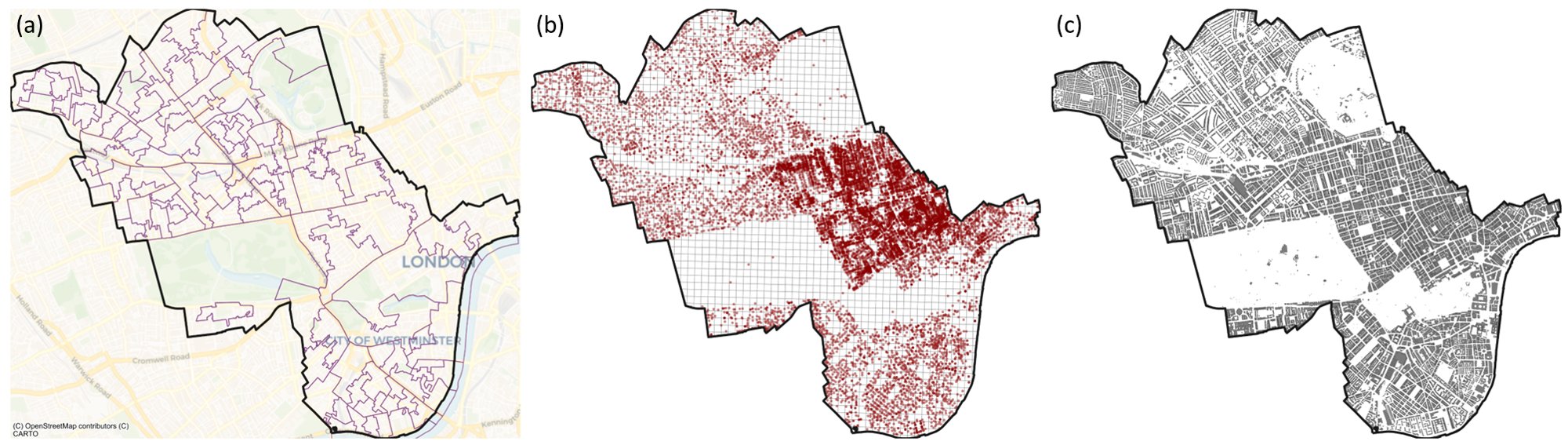}
\caption{GIS layers of the Westminster study area: (a) Westminster boundary overlaid on the base map with Lower Layer Super Output Areas (LSOAs). (b) Westminster boundary with a 100 m grid and postcode distribution. (c) Westminster boundary with building topology.}
\label{fig:1}
\end{figure}

This study focuses on the administrative boundary of Westminster in London,
UK. Located in central London, Westminster covers a total administrative area of 21.5 km$^2$. This study was conducted with support from Westminster City Council. Figure~\ref{fig:1}(a) presents the boundary of the Westminster study area overlaid
on an online basemap, together with the boundaries of Lower Layer Super
Output Areas (LSOAs). Figure~\ref{fig:1}(b) overlays a regular
\(100\ m\  \times \ 100\ m\) grid and postcode point features within the
study area boundary, supporting fine-grained spatial discretisation and
the representation of address distributions. Figure~\ref{fig:1}(c) illustrates the study area boundary
combined with building layers derived from OS MasterMap Topographic Area
data. These polygon-based building and topographic features provide a
direct visual representation of building morphology and the overall
structure of the built environment in Westminster.

\subsection{Data Sources}

Two primary datasets were used to construct the multimodal property
energy performance dataset. The first is the
Energy Performance Certificate (EPC) database \cite{UK_EPC_data}, which provides tabular
and textual inputs for the model. Each property is uniquely
identified by a Unique Property Reference Number (UPRN), and the EPC records include basic building attributes, system-related parameters, and
assessor-recorded textual descriptions. 

The second is the
Property Location and Geometry dataset \cite{OS_MasterMap}, which provides the spatial
input. This dataset uses the Topographic Identifier
(TOID) as the unique identifier for building geometry objects and
includes building footprints, spatial locations, and topological
attributes. In addition, building height information was obtained from the Building Height Attributes (BHA) dataset.

\subsection{Data Pre-processing and Modality Construction}

To align property records with building geometries, EPC records were linked
to spatial objects through UPRN--TOID mapping. The matched geometries were then associated with building polygon features from the OS MasterMap Topographic Area dataset. When a property
corresponds to multiple polygon components, the polygon with the
largest area was selected as the primary building footprint to reduce
interference from ancillary structure.

Based on the linked datasets, three modalities were constructed. The tabular modality was derived from structured EPC attributes and system parameters. The textual modality was built from assessor-recorded property descriptions in the EPC data. The spatial modality combined GIS-based spatial attributes with a geometry-based representation of building footprints. 

To obtain fixed-length and comparable geometric representations,
each building footprint was converted into boundary sequences using equal
arc-length sampling, producing a
two-dimensional point sequence of length \(L = 128\):

\begin{equation}
S = \{(x_{\ell}, y_{\ell})\}_{\ell=1}^{L}.
\end{equation}

The sequence was then normalised by translation and scale. First, the geometric centre of the boundary points was computed
as:

\begin{equation}
\mu = \frac{1}{L}\sum_{\ell=1}^{L} S_{\ell}, 
\qquad 
S'_{\ell} = S_{\ell} - \mu.
\end{equation}

This translation step removes the effect of absolute geographic location and allows the model to focus on shape characteristics. Scale normalisation was then performed using the maximum radial distance from the centre:

\begin{equation}
r_{\max} = \max_{\ell}\|S'_{\ell}\|_{2}, 
\qquad 
\tilde{S}_{\ell} = \frac{S'_{\ell}}{r_{\max} + \varepsilon}.
\end{equation}
where \(\varepsilon\) is a small constant for
numerical stability. This yields a location- scale-invariant boundary
representation, improving training stability and generalisation.

To characterise building orientation, the covariance
matrix of the centred boundary points was constructed as:

\begin{equation}
C = X^{\mathsf{T}} X, 
\qquad 
X = \left[S'_1; \cdots; S'_{\ell}\right].
\end{equation}

The eigenvector \(v\) associated with the largest eigenvalue was taken as the principal axis, and the orientation angle was computed as:

\begin{equation}
\theta = atan2\left( v_{y},v_{x} \right)
\end{equation}

Considering the symmetry of building orientation, the angle was finally
mapped to the range \(\left\lbrack 0,\pi) \right.\ \). After multimodal record linkage, geometry matching, and quality filtering, the final valid dataset comprised 124,990 properties.

\section{Methodology}

\begin{figure}[htbp]
\centering
\includegraphics[width=0.8\linewidth]{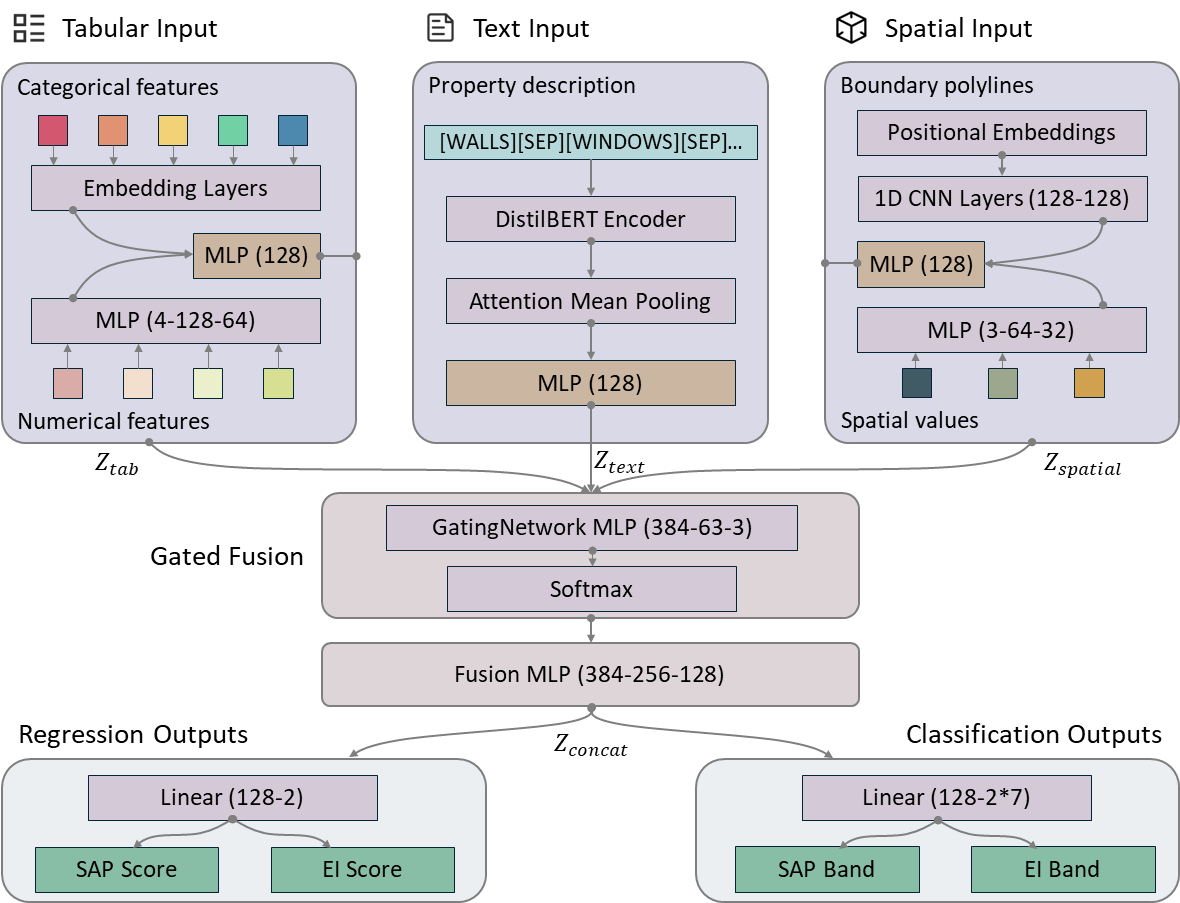}
\caption{Multimodal model framework.}
\label{fig:2}
\end{figure}

To fully exploit multimodal information for property energy performance prediction and to support retrofit decision-making in Westminster, this study proposes a multimodal property energy performance prediction framework, as illustrated in Figure~\ref{fig:2}. The framework is designed for a dual-target regression task, with the objectives of simultaneously predicting two key continuous indicators of building energy performance: the SAP score and the EI score. SAP captures the operational energy efficiency and cost implications of residential buildings, while EI directly reflects associated carbon emissions, making the joint prediction of both metrics critical for aligning property-level retrofit decisions with broader net-zero targets. Reliable, fine-grained estimation of SAP and EI at scale enables policymakers to prioritise interventions, design targeted incentive schemes, and evaluate the system-wide impact of retrofit pathways under constrained public budgets.

The proposed framework adopts an end-to-end workflow that encodes policy-derived property records into modality-specific representations, adaptively fuses them at the property level, and outputs joint continuous predictions for SAP and EI. To capture the heterogeneous determinants of residential energy performance, the framework integrates three complementary modalities: structured tabular variables describing building and system attributes, assessor-recorded free text preserving fine-grained fabric and heating semantics, and spatial information characterising both building geometry and local context. Each modality is processed by a dedicated encoder and projected into a shared latent space.

A gated fusion mechanism then produces sample-specific modality weights and combines the aligned representations into a single fused embedding, allowing the model to emphasise the most informative evidence for each property. This fused embedding is used for the primary dual-target regression task, while an auxiliary band-based classification head provides additional supervision to improve optimisation stability and generalisation.

\subsection{Modality-Specific Encoders}

The detailed feature lists for each modality are summarised in Table~\ref{tab:features}. All selected features are derived from the core parameters of the standardised SAP calculation
procedure, which is detailed in the Supplementary Materials.

\begin{table}[htbp]
\caption{Input features used in the proposed multimodal property energy performance prediction model.}
\label{tab:features}
\centering
\small
\setlength{\tabcolsep}{6pt}
\renewcommand{\arraystretch}{1.2}
\begin{tabularx}{\linewidth}{@{}>{\raggedright\arraybackslash}p{0.35\linewidth} >{\raggedright\arraybackslash}p{0.14\linewidth} >{\raggedright\arraybackslash}X@{}}
\toprule
\textbf{Parameters} & \textbf{Data Type} & \textbf{Description}\\
\midrule
CONSTRUCTION\_AGE\_BAND & Categorical & Age band indicating when the
building (or building part) was constructed. \\
PROPERTY\_TYPE & Categorical & Type of property, e.g. House, Flat,
Maisonette. \\
BUILT\_FORM & Categorical & Building form, e.g. Detached, Semi-detached,
Terraced. \\
ENERGY\_TARIFF & Categorical & Type of electricity tariff applied to the
property. \\
MAIN\_FUEL & Categorical & Primary fuel used for the central heating
system, e.g. gas or electricity. \\
TOTAL\_FLOOR\_AREA & Numerical & Total usable floor area of the
property. \\
NUMBER\_HABITABLE\_ROOMS & Numerical & Number of habitable rooms. \\
NUMBER\_HEATED\_ROOMS & Numerical & Number of rooms served by the
heating system. \\
PHOTO\_SUPPLY & Numerical & Percentage of photovoltaic area relative to
the total roof area. \\
WALLS\_DESCRIPTION & Text & Description of wall construction and
insulation characteristics. \\
WINDOWS\_DESCRIPTION & Text & Description of window type and thermal
performance. \\
FLOOR\_DESCRIPTION & Text & Description of floor construction and
insulation. \\
ROOF\_DESCRIPTION & Text & Description of roof construction and
insulation. \\
MAINHEAT\_DESCRIPTION & Text & Description of the main space heating
system. \\
MAINHEATCONT\_DESCRIPTION & Text & Description of heating control
systems. \\
HOTWATER\_DESCRIPTION & Text & Description of the domestic hot water
system. \\
LIGHTING\_DESCRIPTION & Text & Description of lighting system type and
efficiency. \\
BOUNDARY & GIS (Spatial) & Property footprint boundary sequence. \\
FOOTPRINT\_AREA & GIS (Spatial) & Property footprint area. \\
HEIGHT & GIS (Spatial) & Property height. \\
ORIENTATION & GIS (Spatial) & Property orientation. \\
\bottomrule
\end{tabularx}
\end{table}

For the tabular modality, each categorical feature is encoded using an independent embedding table with embedding dimension \(e\). The resulting embeddings are concatenated as:

\begin{equation}
c = concat\left( E_{1}\left\lbrack x_{1} \right\rbrack,\cdots,E_{M}\left\lbrack x_{M} \right\rbrack \right) \in R^{M \times e}
\end{equation}
where \(M\) denotes the number of categorical fields. Numerical features are encoded using a two-layer multilayer perception (MLP), producing the numerical representation \(n\). The concatenated representation \(\lbrack c,n\rbrack\) is then projected into a shared latent space of dimension \(d\), yielding the tabular embedding:

\begin{equation}
Z_{tab} \in R^{d}
\end{equation}

For the text modality, a Transformer encoder pre-trained on large-scale external corpora is used to encode the concatenated textual sequence, producing token-level hidden representations:

\begin{equation}
H \in R^{T \times h}
\end{equation}
where \(T\) denotes the sequence length and \(h\) the hidden dimension. Mean pooling with attention masks is then applied to obtain a sequence-level representation:

\begin{equation}
P = \frac{\sum_{t = 1}^{T}{m_{t}H_{t}}}{\sum_{t = 1}^{T}m_{t}}
\end{equation}
where \(m_{t}\) denotes the attention mask. The pooled representation is further projected to dimension \(d\), resulting in the text embedding:

\begin{equation}
Z_{text} \in R^{d}
\end{equation}

For the spatial modality, the encoder consists of a boundary sequence branch and a spatial numerical feature branch. For the boundary sequence, each
two-dimensional coordinate point \((x,y)\) is linearly projected into a \(d\)-dimensional embedding and combined with learnable positional encodings, producing:

\begin{equation}
P \in R^{L \times d}
\end{equation}
where \(L\) is the boundary sequence length. Two layers of one-dimensional convolution are then applied to capture local sequential patterns, followed by global average pooling to obtain the boundary representation:

\begin{equation}
s \in R^{d}
\end{equation}

For the spatial numerical features, a three-dimensional feature vector is encoded using an MLP to obtain \(u\). The concatenation \(\lbrack s,u\rbrack\) is then projected to dimension \(d\), producing the spatial embedding:

\begin{equation}
Z_{spatial} \in R^{d}
\end{equation}

\subsection{Gated Fusion Mechanism}

To enable adaptive modality weighting at the sample level, a gated fusion mechanism is introduced. The three modality embeddings are first concatenated as:

\begin{equation}
Z = \left\lbrack Z_{tab};Z_{text};Z_{spatial} \right\rbrack \in R^{3d}
\end{equation}

A two-layer MLP gate then produces modality logits, which are normalised using a softmax function to obtain modality weights:

\begin{equation}
\alpha = softmax\left( g(Z) \right) \in R^{3d},\ \sum_{i}^{}{\alpha_{i} = 1}
\end{equation}

Each modality embedding is rescaled by its corresponding weight and concatenated to form the gated representation:

\begin{equation}
\widetilde{Z} = \left\lbrack \alpha_{tab}Z_{tab};\alpha_{text}Z_{text};\alpha_{spatial}Z_{spatial} \right\rbrack
\end{equation}

The final fused representation \(Z_{fuse}\) is obtained through a fusion MLP. During inference, the modality weights \(\alpha\) can be explicitly extracted for modality contribution analysis and interpretability.

For the main regression task, a linear prediction head jointly estimates SAP and EI:

\begin{equation}
\widehat{y} = WZ_{fuse} + b,\ \ \ \widehat{y} = \left\lbrack {\widehat{y}}_{SAP},{\widehat{y}}_{EI} \right\rbrack \in R^{2}
\end{equation}

In addition, auxiliary band-based classification heads are introduced for each target, outputting seven-class logits:

\begin{equation}
L \in R^{2 \times 7}
\end{equation}

Key architectural settings and hyperparameters are summarised in Table~\ref{tab:arch}.

\begin{table}[htbp]
\centering
\caption{Model architecture and key hyperparameter settings.}
\label{tab:arch}
\small
\setlength{\tabcolsep}{4pt}
\renewcommand{\arraystretch}{1.15}

\begin{tabularx}{\linewidth}{
>{\raggedright\arraybackslash}p{0.19\linewidth}
>{\raggedright\arraybackslash}p{0.22\linewidth}
>{\raggedright\arraybackslash}p{0.20\linewidth}
X}
\toprule
\textbf{Module} & \textbf{Parameter} & \textbf{Setting} & \textbf{Description} \\
\midrule

\multirow{2}{*}{Global} 
& Number of modalities & 3 & Three-modality fusion framework \\
& Unified feature dimension $d$ & 128 & Aligned output dimension of all encoders \\

\midrule
\multirow{5}{*}{Tabular Encoder}
& Categorical embedding dimension & 64 & Independent embedding for each categorical field \\
& Padding / unknown index & 0 & Used for missing or unseen categories \\
& Numerical MLP hidden dims & [128, 64] & Linear + ReLU + Dropout \\
& Dropout rate & 0.1 & Applied to numerical MLP and fusion layers \\
& Output dimension & 128 & Produces $z_{\text{tab}} \in \mathbb{R}^{128}$ \\

\midrule
\multirow{6}{*}{Text Encoder}
& Pre-trained backbone & DistilBERT & Transformer-based text encoder \\
& Field separator token & [SEP] & Concatenation of multiple EPC text fields \\
& Maximum sequence length & 512 & Padding and truncation applied \\
& Pooling strategy & Mask-aware mean pooling & Mean pooling with attention mask \\
& Projection layer & Linear ($h \rightarrow 128$) + ReLU + Dropout (0.1) & $h=768$ for DistilBERT \\
& Backbone fine-tuning & False & End-to-end fine-tuning enabled \\

\midrule
\multirow{7}{*}{Spatial Encoder}
& Boundary sequence length $L$ & 128 & Fixed-length building footprint representation \\
& Boundary point dimension & 2 & 2D planar coordinates \\
& Boundary projection & Linear ($2 \rightarrow 128$) & Maps coordinate to latent space \\
& Positional encoding & Learnable ($128 \times 128$) & Matched to sequence length \\
& Sequence encoder & 2$\times$Conv1D ($128 \rightarrow 128$, kernel=3, padding=1) + ReLU & Extracts local geometric patterns \\
& Global pooling & AdaptiveAvgPool1d & Produces global boundary representation \\
& Spatial numerical MLP & [64, 32] & Linear + ReLU + Dropout (0.1) \\

\midrule
Fusion MLP
& Hidden dims & [256, 128] & Linear + ReLU + Dropout (0.1) \\

\midrule
Regression Head
& Output dimension & 2 & Predicts SAP and EI (continuous values) \\

\midrule
\multirow{2}{*}{Band Auxiliary Head}
& Number of classes & 7 (A--G) & Energy efficiency bands for each target \\
& Output shape & $(B,2,7)$ & Linear ($128 \rightarrow 2\times7$) followed by reshape \\

\bottomrule
\end{tabularx}
\end{table}

\subsection{Prediction Objective and Loss Function}

To achieve robust joint regression of SAP and EI while leveraging discrete efficiency band information as auxiliary supervision, a multi-task objective is adopted. The model is trained to predict the continuous SAP and EI scores, while additional band-based classification tasks are introduced to improve optimisation stability and generalisation.

The regression targets are defined as \(y = \left\lbrack y_{SAP},y_{EI} \right\rbrack\). During training, the target variables are normalised using the mean and standard deviation computed from the training set:

\begin{equation}
y^{norm} = \frac{y - \mu_{y}}{\sigma_{y}}
\end{equation}
where \(\mu_{y}\) and \(\sigma_{y}\) denote the mean and standard
deviation of the target variables in the training set. During validation and testing, both predictions and ground-truth values are transformed back to the original scale for evaluation.

For the main regression task, the Huber loss is adopted to improve robustness to outliers. Given the prediction error \(e = \widehat{y} - y\), the Huber loss is defined as:

\begin{equation}
L_{\text{Huber}}(e) =
\begin{cases}
\frac{1}{2} e^{2}, & |e| \le \delta, \\
\delta |e| - \frac{1}{2}\delta^{2}, & |e| > \delta.
\end{cases}
\end{equation}
where \(\delta\) is the threshold parameter. In practice, the Huber loss is averaged over each training batch.

To incorporate discrete energy efficiency band information as auxiliary supervision, additional band-based classification losses are introduced. Ground-truth band labels are obtained by first transforming the normalised continuous targets back to their original regression scale and then mapping them to the corresponding band indices according to the official SAP and EI band definitions. The detailed score to band mapping is provided in the Supplementary Materials. Cross-entropy losses are computed separately for SAP and EI:

\begin{equation}
L_{band}^{SAP} = CE\left( L_{SAP},b_{SAP} \right),\ \ \ L_{band}^{EI} = CE\left( L_{EI},b_{EI} \right)
\end{equation}
where \(L_{SAP}\) and \(L_{EI}\) denote the predicted band logits, and
\(b_{SAP}\) and \(b_{EI}\) represent the corresponding ground-truth band
labels.

The final loss objective is defined as a weighted combination of the regression loss and the auxiliary band classification losses:

\begin{equation}
L = L_{Huber} + w_{SAP}L_{band}^{SAP} + w_{EI}L_{band}^{EI}
\end{equation}
where \(w_{SAP}\) and \(w_{EI}\) control the relative contribution of the auxiliary classification tasks to the overall objective.

\subsection{Scenario Construction for Retrofit Interventions}

To reduce property-level energy demand and support local decarbonisation, Westminster City Council has prioritised a range of retrofit measures aimed at improving property fabric performance, including wall insulation, roof insulation, and window glazing upgrades. By reducing heat loss from the property envelope, these interventions can lower heating demand and associated energy use, making them an important pathway towards net-zero objectives. To support the identification of priority retrofit areas and the assessment of expected intervention outcomes, this study incorporates a scenario-based retrofit analysis at the property level.

The proposed model predicts continuous SAP and EI scores, which are derived from annual energy cost and equivalent CO$_2$ emissions, respectively. Under SAP 10.2 \cite{BRE_SAP102}, the SAP score is calculated from the annual energy cost, as defined in Eqs.~(\ref{eq:ecf}) and (\ref{eq:sap}):

\begin{equation}
ECF = d \cdot \frac{\mathit{Cost}}{TFA + 45}
\label{eq:ecf}
\end{equation}

\begin{equation}
SAP =
\begin{cases}
108.8 - 120.5 \cdot \log_{10}(ECF), & ECF \geq 3.5 \\
100 - 16.21 \cdot ECF, & ECF < 3.5
\end{cases}
\label{eq:sap}
\end{equation}
where TFA is the total floor area from EPC records and $d$ is a tariff-related deflator treated as constant under a fixed tariff assumption.

Likewise, the EI score is obtained from the annual equivalent CO$_2$ emission, as defined in Eqs.~(\ref{eq:cf}) and (\ref{eq:ei}).

\begin{equation}
CF = \frac{eCO_2}{TFA + 45}
\label{eq:cf}
\end{equation}

\begin{equation}
EI =
\begin{cases}
200 - 95 \cdot \log_{10}(CF), & CF \geq 28.3 \\
100 - 1.34 \cdot CF, & CF < 28.3
\end{cases}
\label{eq:ei}
\end{equation}

For each retrofit scenario, relevant input variables are modified to represent the intervention, and the updated records are fed into the trained model to generate post-retrofit SAP and EI predictions. These predicted scores are then converted back to the implied annual energy cost and equivalent CO$_2$ emissions using the inverse relationships defined by Eqs.~(\ref{eq:ecf})--(\ref{eq:ei}). By comparing the estimated values before and after intervention, the framework enables rapid assessment of the projected impacts of different retrofit strategies and supports the identification of priority retrofit areas at scale.

\section{Model Evaluation}

This section evaluates the proposed multimodal framework from the perspectives of predictive accuracy, robustness, and optimisation behaviour. Following the methodological design introduced in Section 3, we first describe the dataset splitting strategy and training configuration, and then assess model performance through overall prediction results, modality ablation experiments, and subgroup analyses across different property categories.

\subsection{Dataset Splitting and Evaluation Setup}

To ensure consistent distributions of property types and energy
performance indicators across the training, validation, and test sets, a joint binning and stratified sampling strategy
is adopted for dataset splitting. Let the full dataset be denoted
as:

\begin{equation}
D = \left\{ x_{i} \right\}_{i = 1}^{N}
\end{equation}
where each sample \(x_{i}\) is associated with a property type \(P_{i}\), a
Standard Assessment Procedure score \(S_{i}(SAP)\), and an Environmental
Impact score \(E_{i}(EI)\). These continuous SAP and EI scores are further mapped to their corresponding EPC efficiency bands, denoted \(S_{i}^{(g)}\) and \(E_{i}^{(g)}\), respectively. A joint stratification label is then constructed for each sample using the triplet of
property type, SAP band, and EI band:

\begin{equation}
H_{i} = \left( P_{i},S_{i}^{(g)},E_{i}^{(g)} \right)
\end{equation}

\begin{figure}[htbp]
\centering
\includegraphics[width=\textwidth]{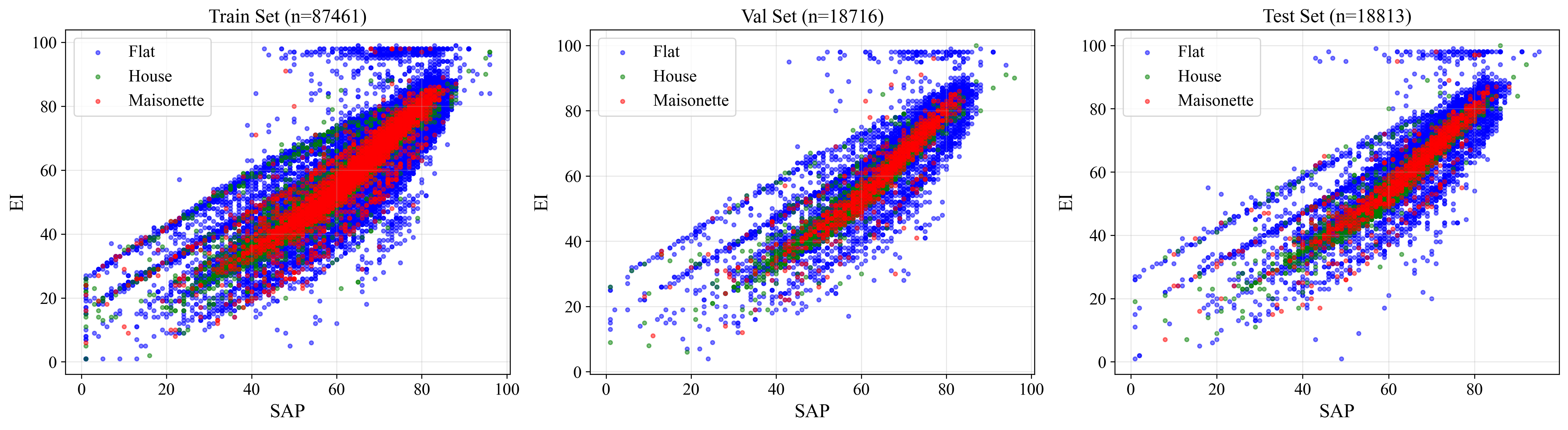}
\caption{Jointly stratified training, validation, and test sets based on property type, SAP band, and EI band.}
\label{fig:4}
\end{figure}

This joint binning scheme simultaneously constrains the distributions of property type, SAP bands, EI bands, and the joint distribution of SAP and EI, thereby improving distributional consistency across data subsets. Based on these joint strata, samples within each stratum are randomly
shuffled and split into training, validation, and test sets using a 70\%
/ 15\% / 15\% ratio. For the \(k\)-th joint stratum containing \(n_{k}\)
samples, the number of samples allocated to each subset is defined as:

\begin{equation}
n_{k}^{train} = \left\lfloor n_{k} \times r_{train} \right\rfloor,\ \ \ n_{k}^{val} = \left\lfloor n_{k} \times r_{val} \right\rfloor,\ \ \ n_{k}^{test} = n_{k} - n_{k}^{train} - n_{k}^{val}
\end{equation}
where \(r_{train} = 0.7\), \(r_{val} = 0.15\), and \(r_{test} = 0.15\). The final training, validation, and test sets are obtained by
aggregating samples from all joint strata. This strategy reduces the risk that certain property types or efficiency bands are under-represented or absent in the validation and test sets, and helps mitigate severe distributional bias caused by limited sample sizes. The resulting stratification outcomes are illustrated in Figure~\ref{fig:4}.

The model is trained end-to-end using the Adam optimiser \cite{kingma2014adam} with layer-wise
learning rates to accommodate heterogeneous initialisation scales: the
pre-trained Transformer backbone uses $1 \times 10^{-5}$, its projection
layers use $1 \times 10^{-4}$, and all remaining modules use
$1 \times 10^{-3}$. A validation-loss-based scheduler is applied to halve the
learning rate when no improvement is observed for five consecutive
epochs. The batch size is set to 128, gradient norms are clipped to a
maximum of 1, and training runs for up to 50 epochs with early stopping
triggered after 10 epochs without validation improvement.

To mitigate bias from imbalanced energy efficiency distributions, a
partition-balanced batch sampling strategy is adopted. The seven EPC
bands (A--G) are merged into five partitions---[AB], [C], [D], [E],
[FG]---and each batch is constructed to reflect the overall partition
proportions for both SAP and EI, ensuring consistent representation of
high, medium, and low efficiency ranges. Detailed justification for the
five-partition design and the full sampling procedure are provided in the
Supplementary Materials.

\subsection{Training Dynamics and Convergence Analysis}

Let the validation set contain \(n\) samples. For a target variable
\(t\) (SAP or EI), the ground-truth values and predictions are denoted
by \(\left\{ y_{i,t} \right\}_{i = 1}^{n}\) and
\(\left\{ {\widehat{y}}_{i,t} \right\}_{i = 1}^{n}\), respectively. All
values are de-normalised to the original scale prior to evaluation.
Three standard regression metrics are used to assess model performance.

The Mean Absolute Error (MAE) measures the average magnitude of
prediction errors and shares the same unit as the target variable:

\begin{equation}
{MAE}_{t} = \frac{1}{n}\sum_{i = 1}^{n}\left| y_{i,t} - {\widehat{y}}_{i,t} \right|
\end{equation}

The Root Mean Squared Error (RMSE) penalises larger errors more strongly
and reflects sensitivity to extreme deviations:

\begin{equation}
{RMSE}_{t} = \sqrt{\frac{1}{n}\sum_{i = 1}^{n}\left( y_{i,t} - {\widehat{y}}_{i,t} \right)^{2}}
\end{equation}

The coefficient of determination \(R^{2}\) quantifies the proportion of
variance explained by the model:

\begin{equation}
R_{t}^{2} = 1 - \frac{\sum_{i = 1}^{n}\left( y_{i,t} - {\widehat{y}}_{i,t} \right)^{2}}{\sum_{i = 1}^{n}\left( y_{i,t} - {\overline{y}}_{t} \right)^{2}},\ \ \ {\overline{y}}_{t} = \frac{1}{n}\sum_{i = 1}^{n}y_{i,t}
\end{equation}

For the dual-target regression task, an overall error indicator is
defined as the arithmetic mean of the MAE values for SAP and EI:

\begin{equation}
Mean_{MAE} = \frac{{MAE}_{SAP} + {MAE}_{EI}}{2}
\end{equation}

Model optimisation is performed in the normalised space,
\(\widetilde{y} = \frac{(y - \mu)}{\sigma}\), to improve numerical
stability and ensure consistent scaling across targets. All evaluation
metrics are computed after de-normalisation,
\(\widehat{y} = \widetilde{y}\sigma + \mu\), to preserve physical
interpretability.

\begin{figure}[htbp]
\centering
\includegraphics[width=.8\textwidth]{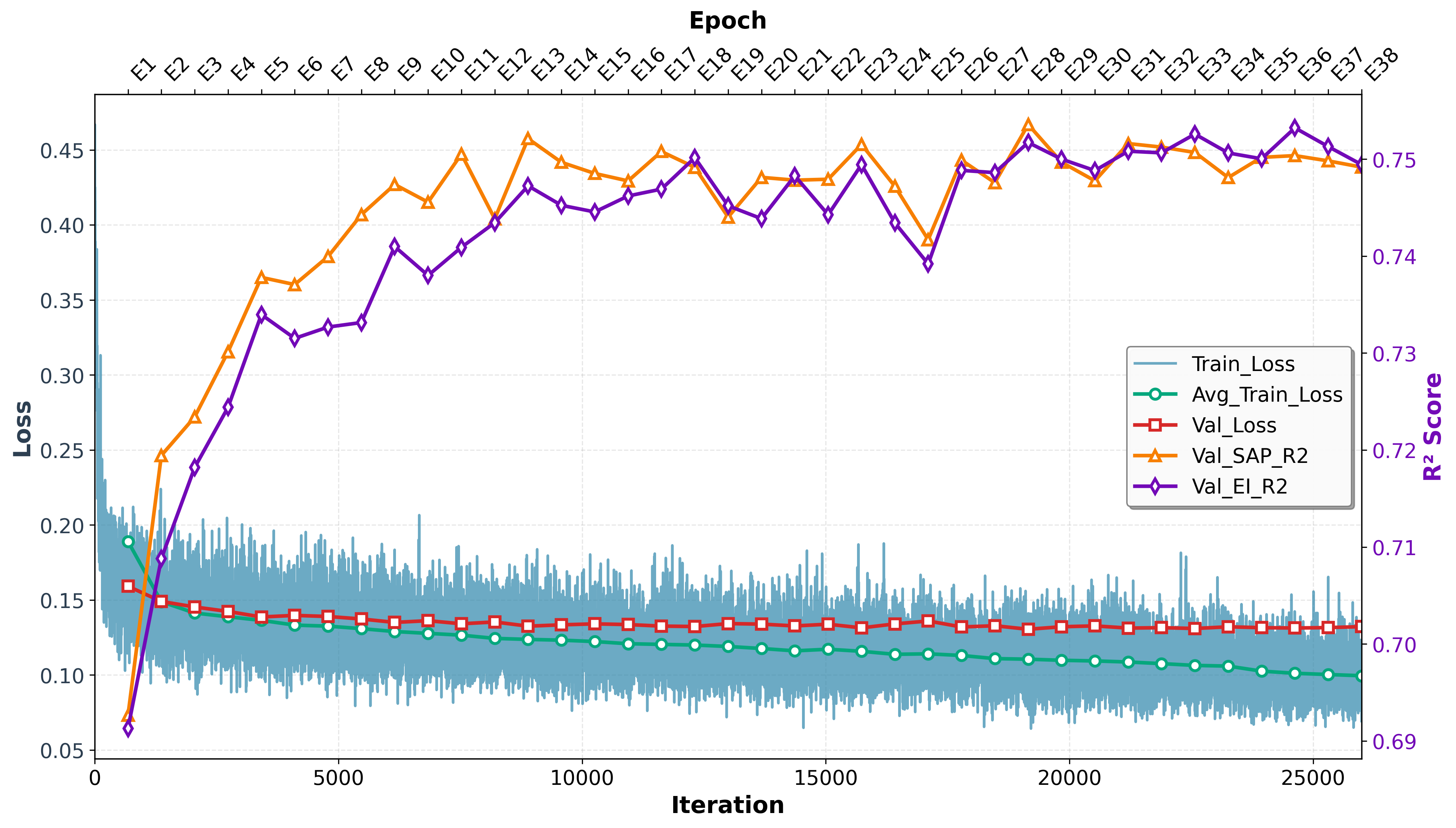}
\caption{Training loss and validation metrics.}
\label{fig:5}
\end{figure}

The training process triggered the early stopping criterion, and
optimisation was terminated at epoch 38. The evolution of training loss
and validation metrics is illustrated in Figure~\ref{fig:5}. The training curves
show a rapid reduction in both training loss and validation errors
during the early epochs, followed by gradual convergence. No significant
divergence between training and validation performance is observed,
indicating stable optimisation and effective regularisation in the
multimodal learning framework.

On the test set, the proposed model demonstrates robust predictive
performance for both SAP and EI. Specifically, for SAP prediction, the
model achieves an MAE of 4.033, an RMSE of 5.739, and an \(R^{2}\) value
of 0.757, while for EI prediction, the corresponding MAE, RMSE, and
\(R^{2}\) are 4.756, 6.711, and 0.748, respectively. By averaging the
MAE values of SAP and EI, an overall dual-target error of
\(Mean_{MAE} = 4.394\) is obtained. These results indicate that the
proposed multimodal model has successfully captured the dominant factors
influencing building energy performance and achieves robust and
consistent predictive accuracy for both SAP and EI.

\subsection{Multimodal Ablation Study}

To systematically evaluate the contribution of each modality in the
proposed multimodal framework, a modality ablation study was conducted
on the test set. Seven model configurations were evaluated, including
three single-modality models (Tabular, Text, and Spatial), three
dual-modality models (Tabular+Text, Tabular+Spatial, and Text+Spatial),
and the full multimodal model (Tabular+Text+Spatial). All models were
trained and evaluated under identical data splits and training settings
to ensure fair comparison.

Model performance was assessed from both continuous regression and
discrete energy band perspectives. For continuous SAP and EI prediction,
the coefficient of determination (\(R^{2})\) was used to measure
explanatory power. For discrete evaluation, Band Accuracy was adopted as
the classification metric. To mitigate instability caused by sparse
samples in extreme efficiency bands, the original seven EPC bands (A--G)
were merged into five groups: {[}AB{]}, {[}C{]}, {[}D{]}, {[}E{]}, and
{[}FG{]}, and accuracy was computed in this merged label space as:

\begin{equation}
{Acc}_{band} = \frac{1}{n}\sum_{i = 1}^{n}{I\left( {\widehat{b}}_{i} = b_{i} \right)}
\end{equation}
where \({\widehat{b}}_{i}\) and \(b_{i}\) denote the predicted and
ground-truth band labels for sample \(i\), respectively, with
\({\widehat{b}}_{i},\ b_{i} \in \left\{ AB,C,D,E,FG \right\}\), and
\(I( \bullet )\) is the indicator function.

\begin{figure}[htbp]
\centering
\includegraphics[width=1.0\linewidth]{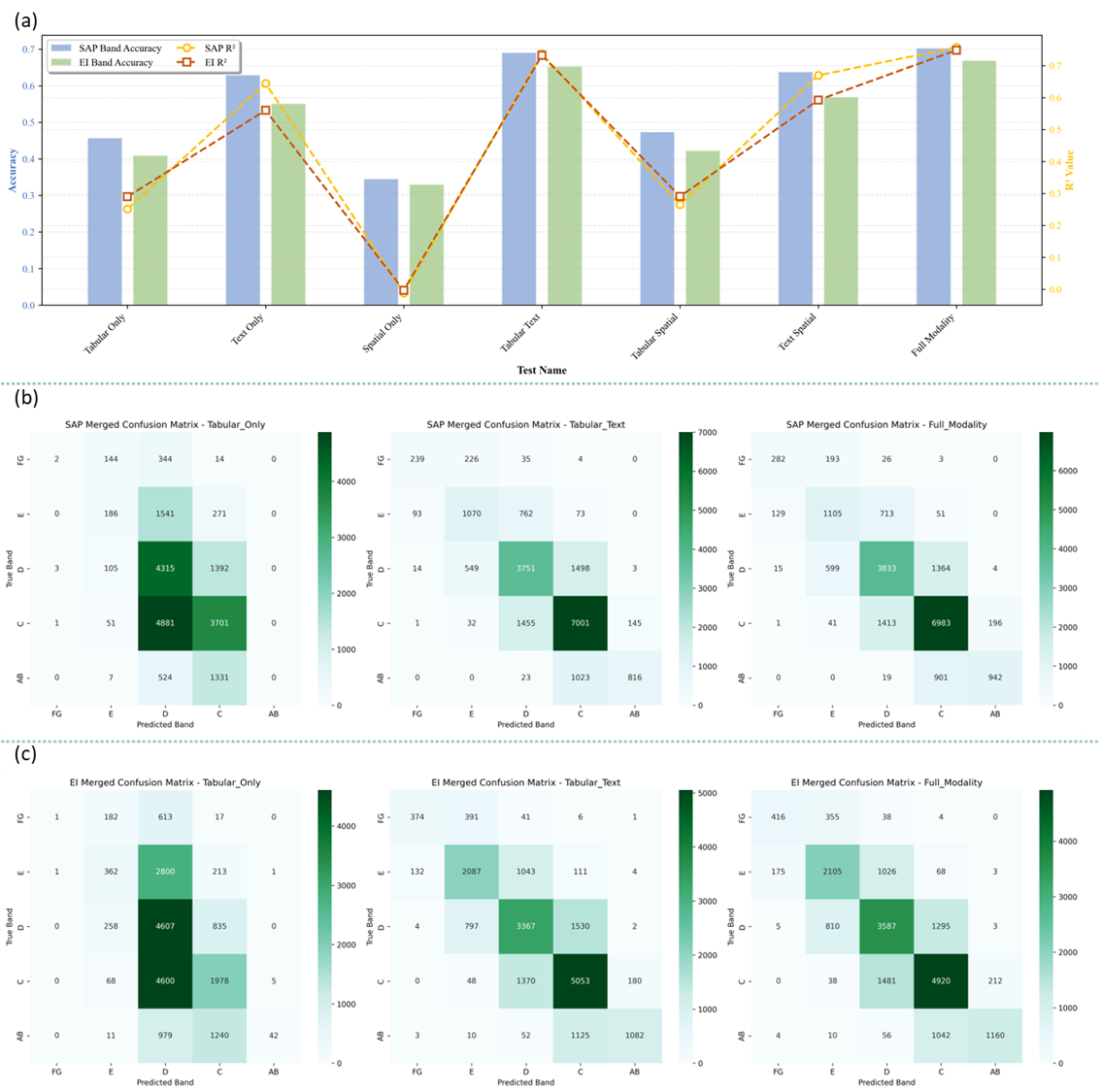}
\caption{Results of modality ablation: (a) Modality comparison in terms of band classification accuracy and R$^2$ values. (b) SAP band prediction results using Tabular, Tabular + Text, and Full Modality inputs. (c) EI band prediction results using Tabular, Tabular + Text, and Full Modality inputs.}
\label{fig:7}
\end{figure}

As shown in Figure~\ref{fig:7}, the full multimodal model consistently achieves
the best performance in terms of both continuous prediction (\(R^{2})\)
and band classification accuracy for SAP and EI. Compared with the
Tabular+Text model, which yields the second-best performance, the full
model improves \(R^{2}\) by 2.4\% for SAP and 2.9\% for EI and increases
Band Accuracy by 2.2\% and 1.9\% for SAP and EI, respectively. These
improvements indicate that spatial information provides complementary
value beyond tabular and textual features. Confusion matrices for all seven modality
configurations are provided in the Supplementary Materials.

\subsection{Performance across Different Subgroups}

To assess the robustness of the proposed framework across heterogeneous property categories, subgroup analyses were conducted for property types, built forms, and construction age bands.

\begin{figure}[htbp]
\centering
\includegraphics[width=0.9\linewidth]{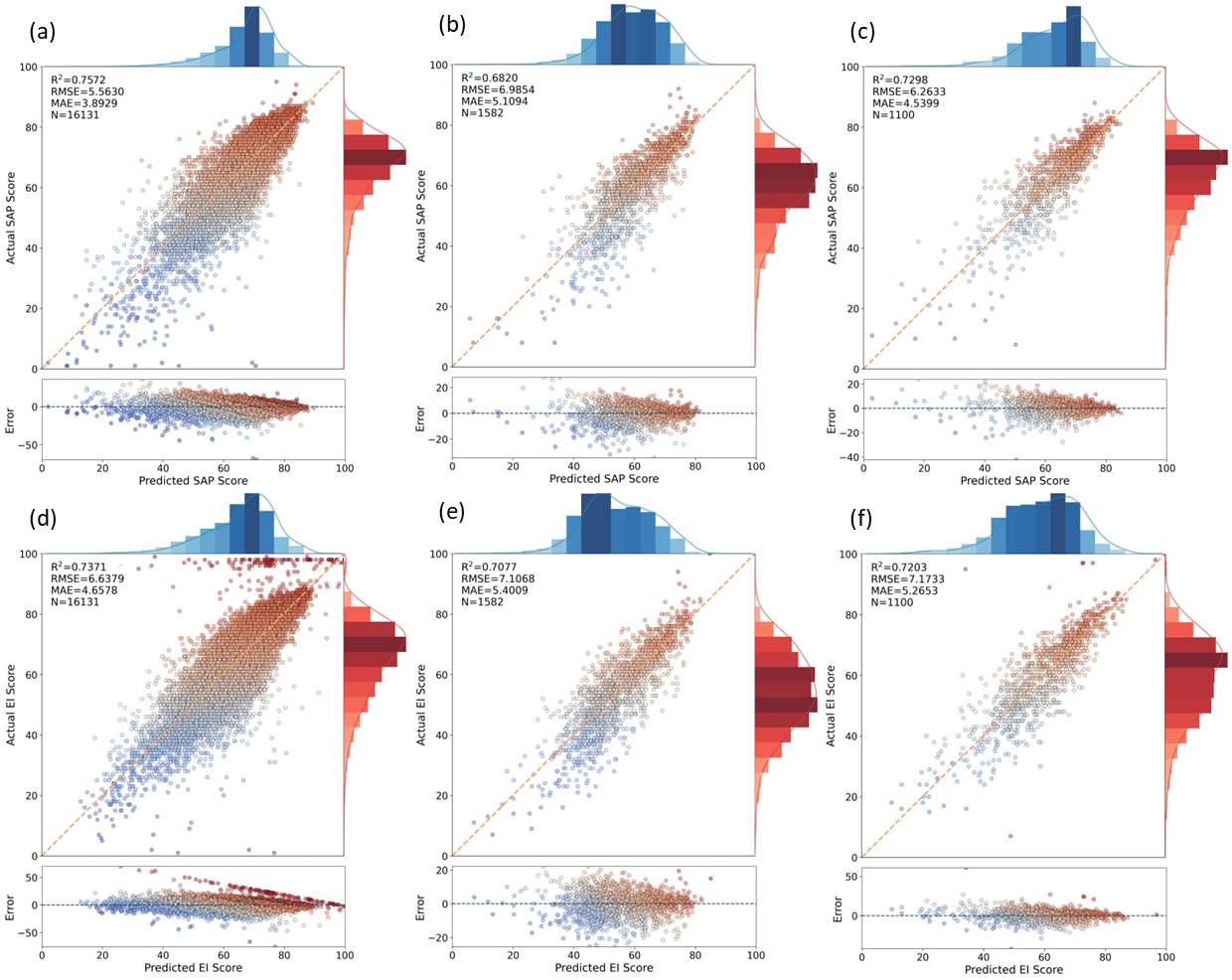}
\caption{Predicted versus actual SAP and EI regression results by property type: (a--c) SAP predictions for Flats, Houses, and Maisonettes. (d--f) EI predictions for Flats, Houses, and Maisonettes.}
\label{fig:8}
\end{figure}

Figure~\ref{fig:8} presents a subgroup analysis of predictive performance
across property types. The model achieves the best performance for Flats
in both SAP and EI prediction, while the weakest performance is observed
for Houses. This pattern is consistent across both targets, reflecting
the strong correlation between SAP and EI. Larger performance discrepancies are observed for SAP than
for EI: the maximum $R^{2}$ difference across property types reaches
0.0752 for SAP versus 0.0294 for EI, and the maximum MAE difference is
1.2165 for SAP versus 0.7431 for EI. Houses exhibit substantially
greater internal heterogeneity than Flats and Maisonettes in
construction age, size, fabric, and heating configurations, which likely
underlies this pattern. Despite these differences, the model provides
stable predictions for all subgroups without systematic failure or
extreme bias.

\begin{figure}[htbp]
\centering
\includegraphics[width=\textwidth]{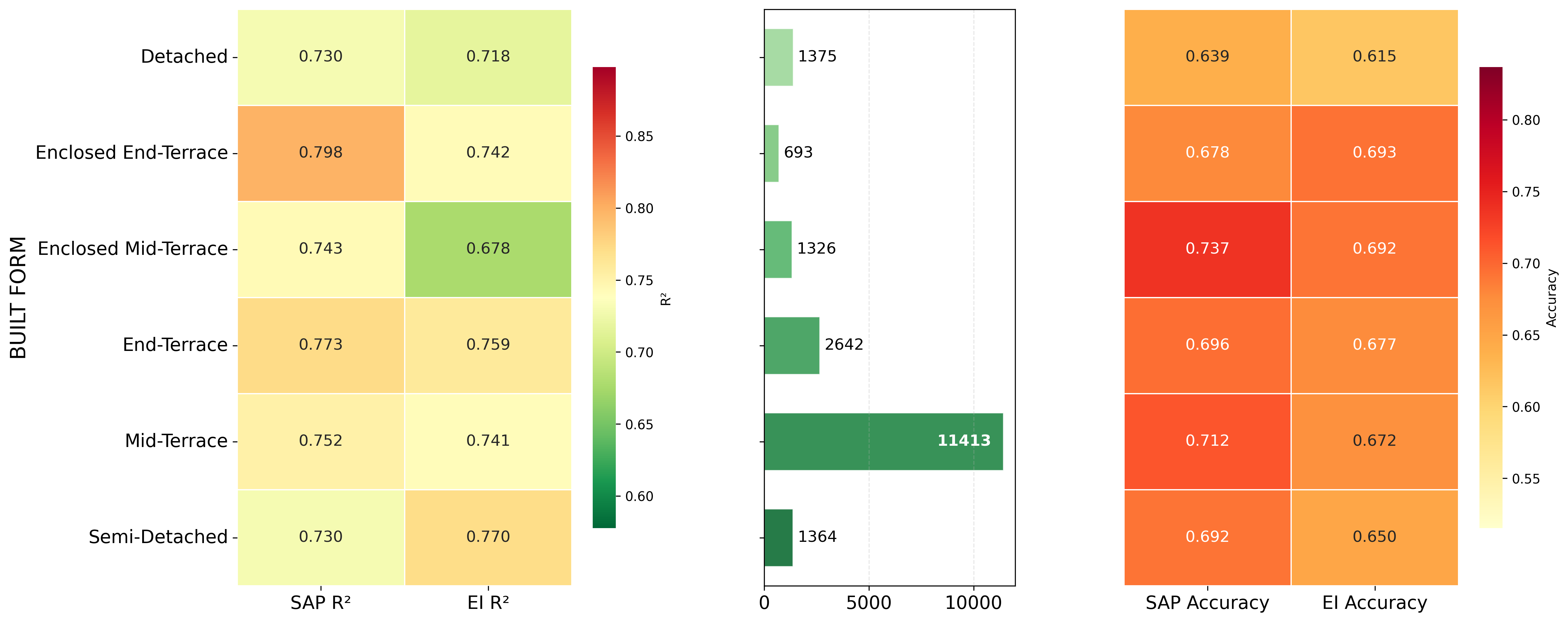}
\caption{Prediction results of SAP and EI scores and bands across different built forms: the left figure shows the R\textsuperscript{2} for SAP and EI score prediction; the middle figure reports the sample counts for each built form; the right figure illustrates the prediction accuracy for SAP and E bands.}
\label{fig:9}
\end{figure}

Figure~\ref{fig:9} presents a subgroup analysis across built forms. The
$R^{2}$ values for SAP and EI remain consistent across categories, with
no substantial degradation for any specific built form. The highest SAP
prediction performance is observed for Enclosed End-Terrace properties,
while Semi-Detached properties yield the best EI prediction. Notably,
the relative performance ranking based on continuous scores does not
always align with band classification accuracy. This discrepancy arises
because modest score errors may still fall within the correct band
interval, further highlighting the importance of fine-grained continuous
prediction for retrofit planning where decisions depend on subtle
performance differences.

\begin{figure}[htbp]
\centering
\includegraphics[width=\textwidth]{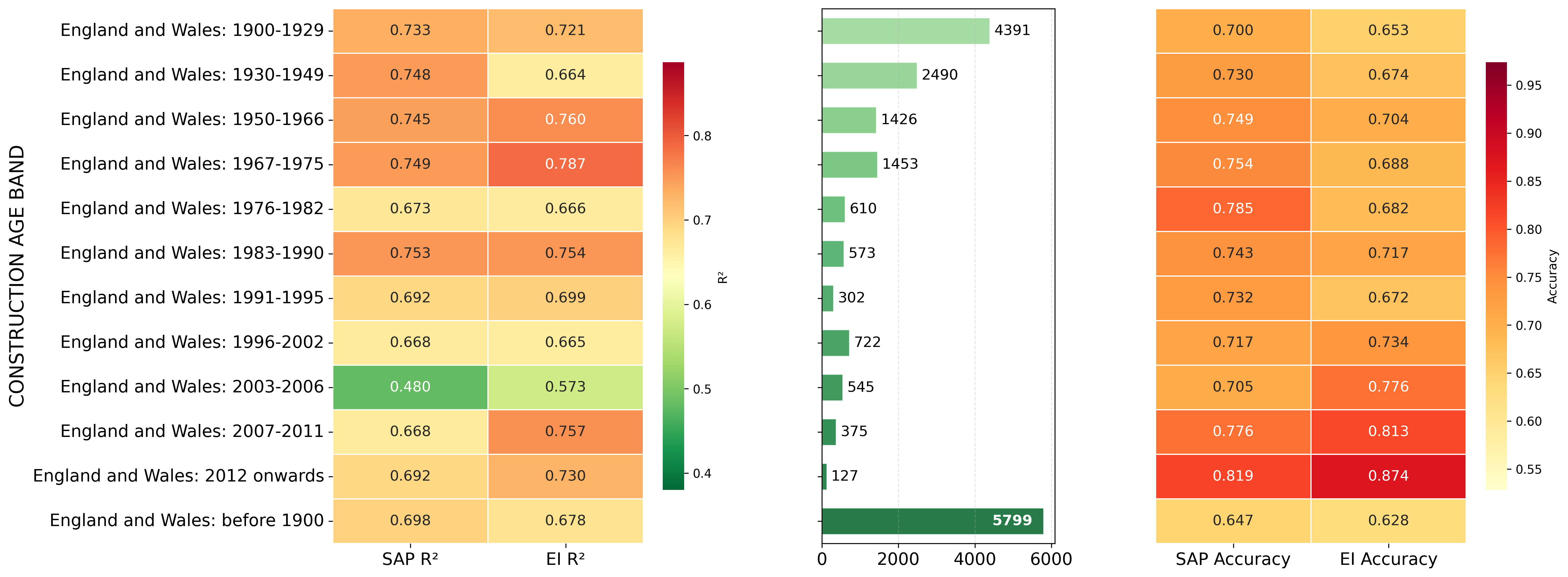}
\caption{Prediction results of SAP and EI scores and bands across construction age band: the left figure shows the R\textsuperscript{2} for SAP and EI score prediction; the middle figure reports the sample counts for each construction age band; the right figure illustrates the prediction accuracy for SAP and E bands.}
\label{fig:10}
\end{figure}

Figure~\ref{fig:10} presents a subgroup analysis across construction age
bands. The model performs well across all periods, with particularly
strong band accuracy (exceeding 0.8) for properties built from 2012
onwards. However, for properties built between 2003 and 2006, $R^{2}$
values for both SAP and EI drop to around 0.5. A closer examination
reveals that scores for this age band are concentrated within the 60--90
range with a high proportion of upper-segment samples, leading to larger
errors at the extremes of the efficiency spectrum. Despite this
score-level reduction, band accuracy remains above 0.7, and detailed
results are provided in the Supplementary Materials. All other
construction age groups show strong prediction performance with no
evidence of systematic degradation.

Taken together, these findings indicate that the proposed framework maintains robust predictive performance across diverse property subgroups, with no evidence of systematic degradation.

\section{Interpretability and Modality Attribution Analysis}

Having established the predictive validity of the proposed framework, this section investigates how different modalities and features contribute to the model’s predictions. Interpretability analysis is conducted at multiple levels, including sample-wise gated fusion weights, tabular feature attribution, field-level textual importance, and spatial contribution analysis.

\subsection{Distribution of Gated Fusion Weights}

\begin{figure}[htbp]
\centering
\includegraphics[width=0.5\linewidth]{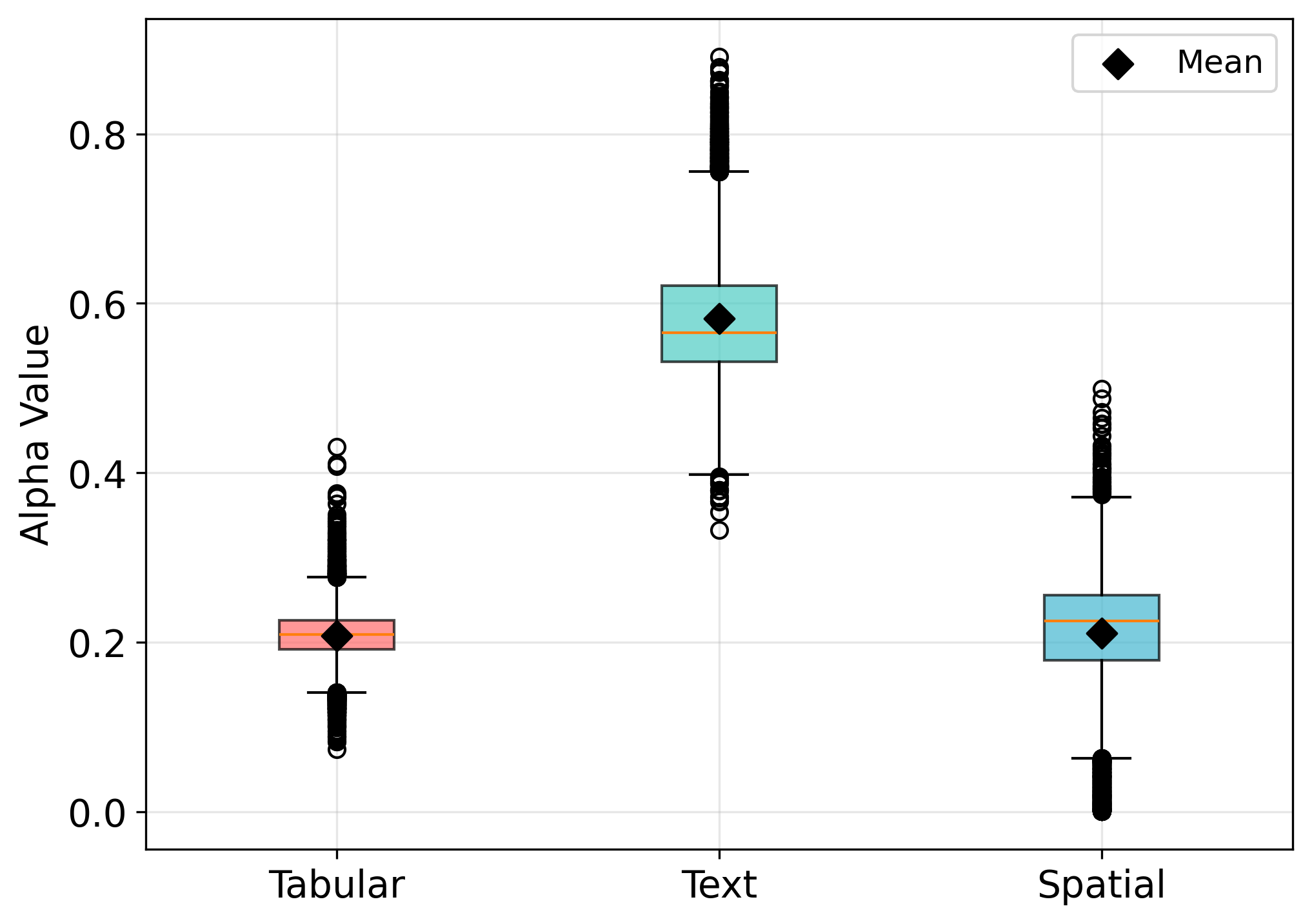}
\caption{Gated fusion alpha value distribution.}
\label{fig:6}
\end{figure}

Figure~\ref{fig:6} presents the distribution of sample-wise fusion weights
produced by the gated fusion mechanism. Overall, the model relies
predominantly on the text modality when predicting SAP and EI, followed
by the spatial modality, while the tabular modality contributes the
least. The dominance of the text modality can be attributed to the
nature of the information it encodes. EPC textual fields provide
fine-grained descriptions of heating systems, domestic hot water
systems, control strategies, and building fabric characteristics such as
walls, windows, and roofs. These elements directly determine the
computational pathways and outcomes of SAP and EI assessments. In
contrast, tabular features typically offer coarse categorical
abstractions of similar information, which may obscure important
variations captured in assessor-written text, resulting in a lower but
consistent contribution to the prediction.

An examination of weight variability further reveals distinct modality
roles. Although the text modality exhibits the largest absolute standard
deviation, its relative variability is moderate given its high mean
weight, indicating that textual information consistently serves as a
primary information source across most samples. The spatial modality
shows the largest relative variability, suggesting strong sample
dependency: geometric and spatial characteristics are highly informative
for certain properties, while being less influential for others. By
comparison, the tabular modality displays the most concentrated weight
distribution, acting as a stable baseline that contributes modestly but
consistently across the dataset, without dominating individual
predictions.

The gated fusion weight distributions highlight a clear
functional separation among modalities. Textual information serves as
the main driver of energy performance prediction, spatial features act
as a property-specific auxiliary signal, and tabular attributes provide
a steady foundational context. These findings empirically support the
necessity of multimodal modelling and offer insights into the relative
importance of different data sources for EPC-based energy performance
assessment and retrofit analysis.

\subsection{Tabular Feature Importance via SHAP}

To quantify the marginal contributions of tabular features to model
predictions, SHAP (SHapley Additive exPlanations) \cite{lundberg2017unified} is employed for
feature attribution, with the detailed methodology provided in the Supplementary Materials. 

\begin{figure}[htbp]
\centering
\includegraphics[width=0.6\linewidth]{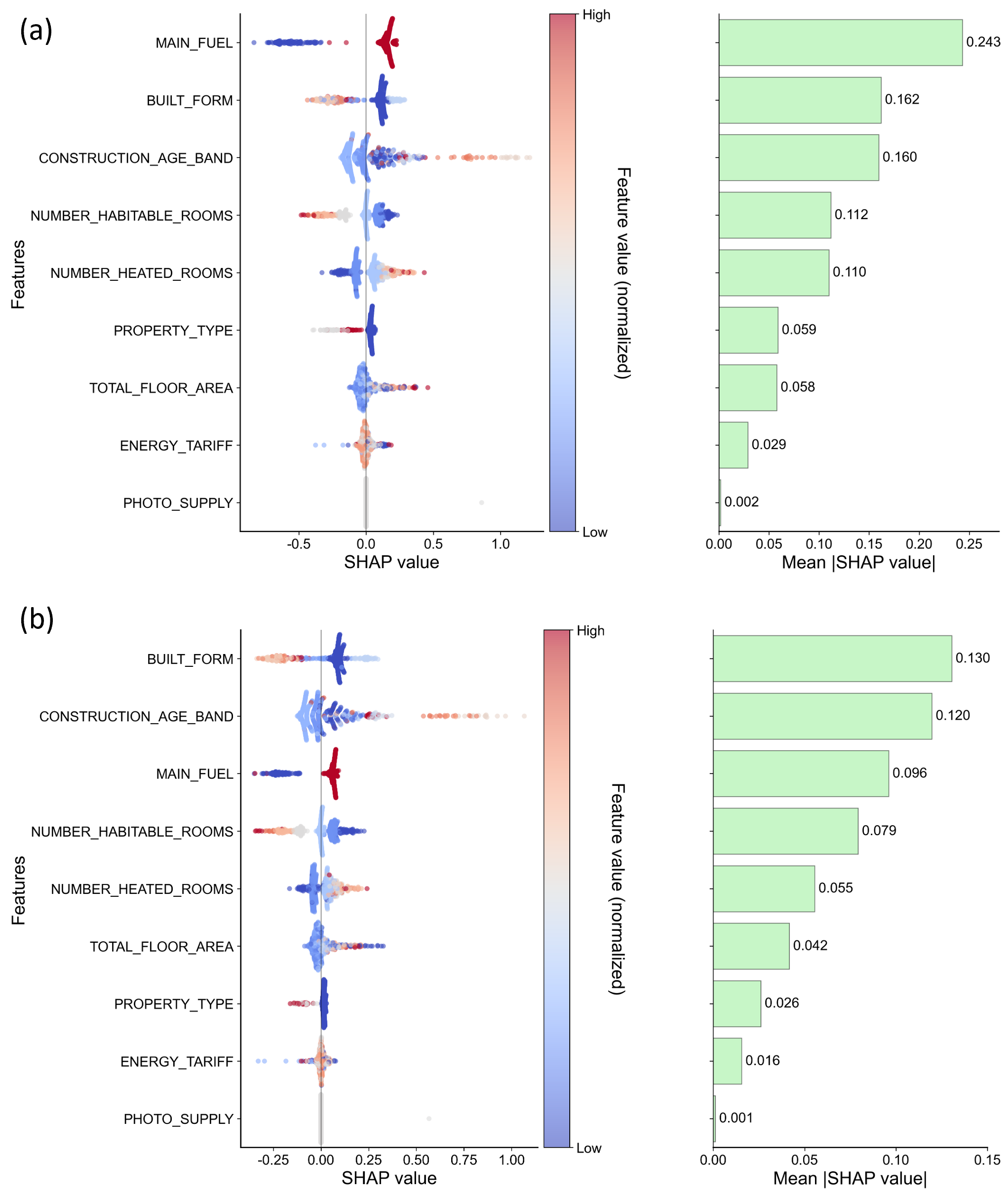}
\caption{Tabular feature importance analysis: (a) SAP; (b) EI.}
\label{fig:11}
\end{figure}

A global interpretability analysis of tabular features was
conducted on the test set to quantify their overall contributions to SAP
and EI predictions and to provide insights into the model's
decision-making process. Figure~\ref{fig:11} presents the SHAP-based feature
importance results, where importance is measured as the mean absolute
SHAP value across test samples. Overall, the three most influential
features for both SAP and EI prediction are main fuel, built form, and
construction age band, indicating that fuel type, building morphology,
and construction period are fundamental structural determinants of
building energy performance. However, although these features dominate
in both tasks, their relative importance rankings differ substantially
between SAP and EI. This divergence highlights that the influence of
tabular features is target-dependent, reflecting differences in the
underlying definitions and calculation logic of SAP and EI.

Notably, the PV supply feature exhibits near-zero global importance for
both SAP and EI. This does not imply that photovoltaic systems are
irrelevant to building energy performance; rather, it reflects
limitations in the EPC data. In the EPC dataset, PV supply values
typically take discrete levels such as 0, 50\%, and 100\%, resulting in
a non-continuous distribution, and most properties are recorded with a
value of zero. This pattern is largely attributable to delayed updates
in EPC records. Although residential photovoltaic installations have
increased rapidly in the UK in recent years, these changes are not yet
fully captured in EPC data. As a result, during SHAP-based importance
analysis, such high-frequency, low-information discrete features are
effectively treated as noise and consequently receive negligible
importance values at the global level.

\subsection{Textual Feature Importance Analysis}

For the text modality, a field-level [MASK] occlusion analysis is
adopted to quantify the contribution of each EPC textual field to the
model's predictions. For each sample, the baseline prediction is first
obtained using the complete text input. A given text field is then
replaced entirely with the [MASK] token, and the importance of that
field is defined as the average absolute change in prediction across the
test set:

\begin{equation}
I_{k,t} = \frac{1}{N}\sum_{n = 1}^{N}\left| {\widehat{y}}_{n,t} - {\widehat{y}}_{n,t}^{( - k)} \right|
\label{eq:text_imp}
\end{equation}
where \({\widehat{y}}_{n,t}\) denotes the original prediction for target \(t\), and \({\widehat{y}}_{n,t}^{( - k)}\) denotes the prediction after masking the \(k\)-th text field.

\begin{figure}[htbp]
\centering
\includegraphics[width=0.8\linewidth]{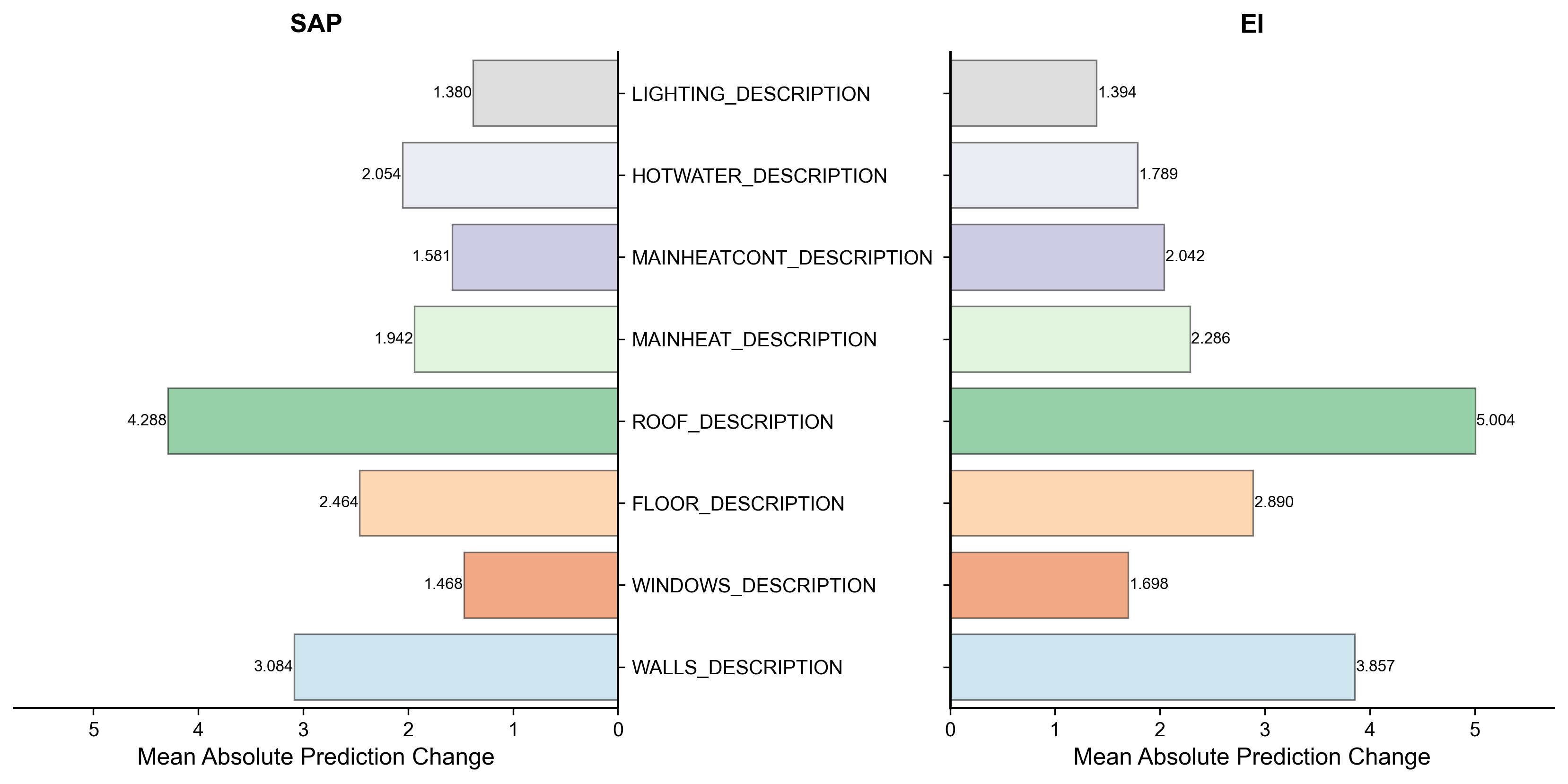}
\caption{Text field importance.}
\label{fig:13}
\end{figure}

Figure~\ref{fig:13} presents the overall results. For both SAP and EI,
roof description emerges as the most influential textual feature,
followed by wall description, while heating system description
contributes the most among system-related fields. From a physical
perspective, fabric-related elements such as roofs and walls directly
govern heating demand, energy consumption, and associated costs and
emissions.

\subsection{Spatial Feature Attribution Analysis}

For the interpretability analysis of the spatial modality, two distinct
sources of information are considered: spatial numerical features and
building boundary sequences. For spatial numerical features, permutation
importance is employed by randomly shuffling individual features and
measuring the resulting change in model predictions, thereby quantifying
their contribution to SAP and EI estimation. For the boundary sequence,
which represents structured geometric information, a sample-level
boundary permutation strategy is adopted, where building boundary
sequences are exchanged across samples while keeping other modalities
and spatial numerical features unchanged. This approach enables an
assessment of the overall impact of building geometry on model
predictions. By disentangling and separately analysing these two
components, the proposed framework provides a systematic and
interpretable understanding of how spatial information contributes to
multimodal energy performance prediction, with detailed methodological descriptions provided in the Supplementary Materials.

\subsubsection{Spatial Numerical Feature Importance}

\begin{figure}[htbp]
\centering
\includegraphics[width=0.6\linewidth]{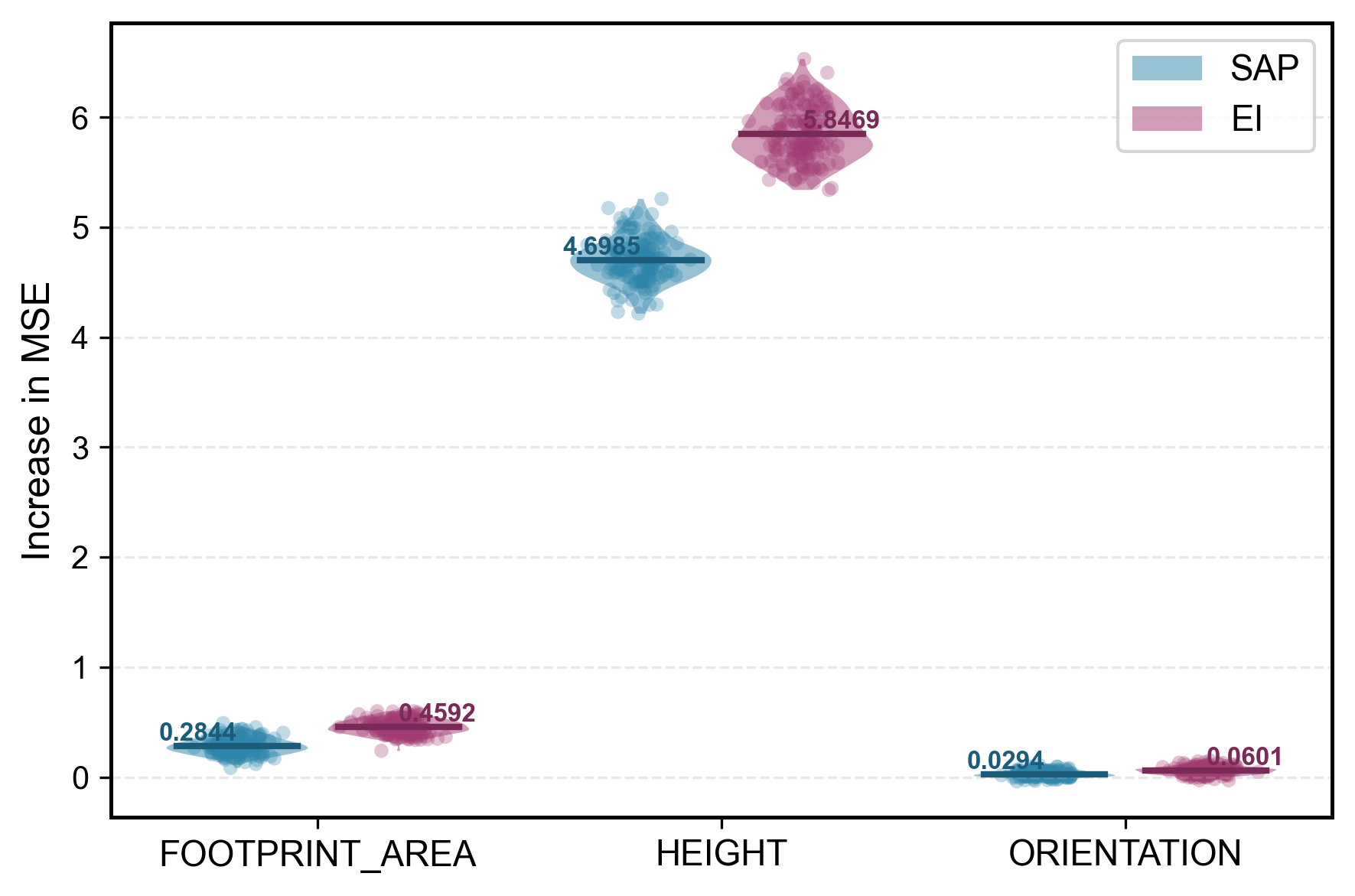}
\caption{Spatial numerical feature importance.}
\label{fig:16}
\end{figure}

Figure~\ref{fig:16} presents the permutation importance of spatial
numerical features. Height emerges as the most influential feature for
both SAP and EI, followed by footprint area, while orientation exhibits
a minor contribution. Height and footprint area jointly determine
building volume, which directly influences heating demand and energy
consumption, whereas the marginal impact of solar gains via orientation
is comparatively small.

\subsubsection{Spatial Geometry Feature Importance}

\begin{figure}[htbp]
\centering
\includegraphics[width=\linewidth]{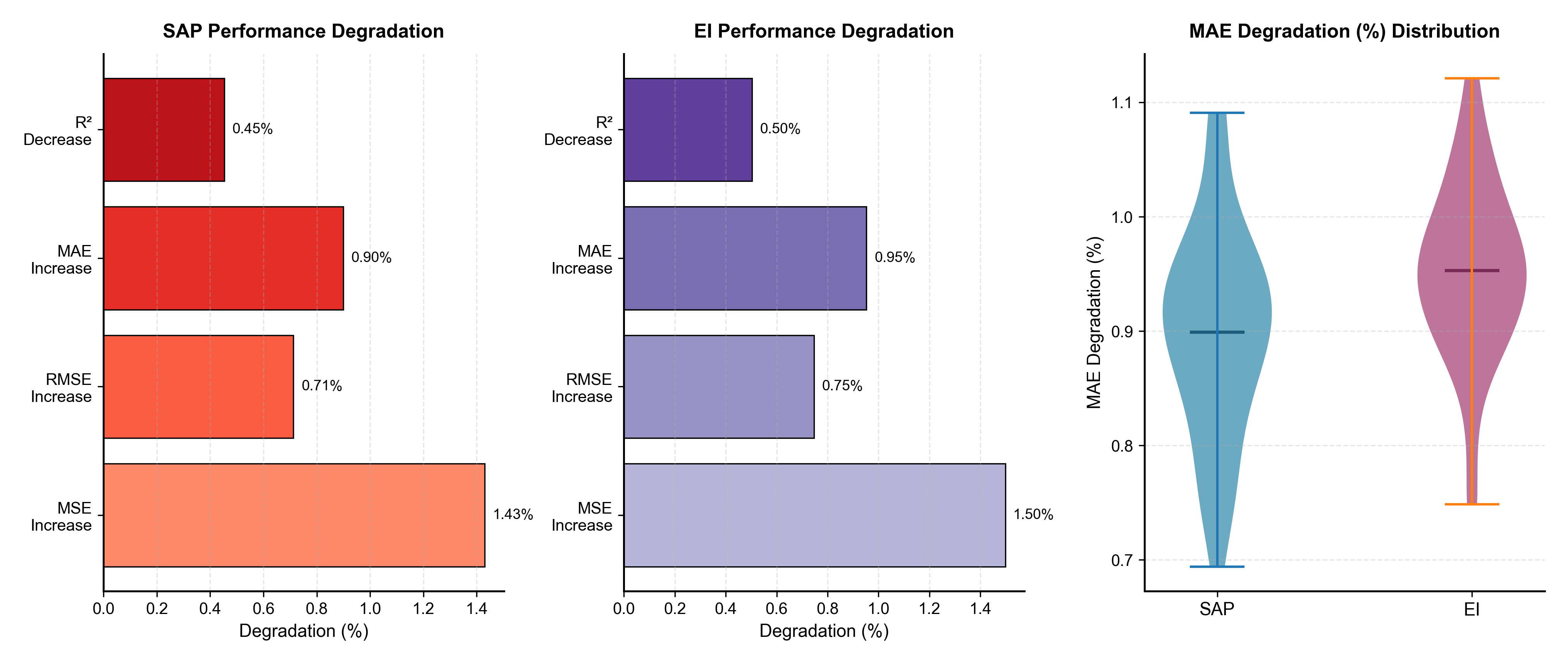}
\caption{Spatial boundary shape information importance analysis.}
\label{fig:19}
\end{figure}

Figure~\ref{fig:19} presents the global boundary permutation results.
When boundary sequences are permuted across samples, prediction MAE for
both SAP and EI increases by approximately 1\%, with corresponding
degradations in $R^{2}$ and RMSE, confirming that boundary geometry
provides complementary information for energy performance prediction.

\begin{figure}[htbp]
\centering
\includegraphics[width=\linewidth]{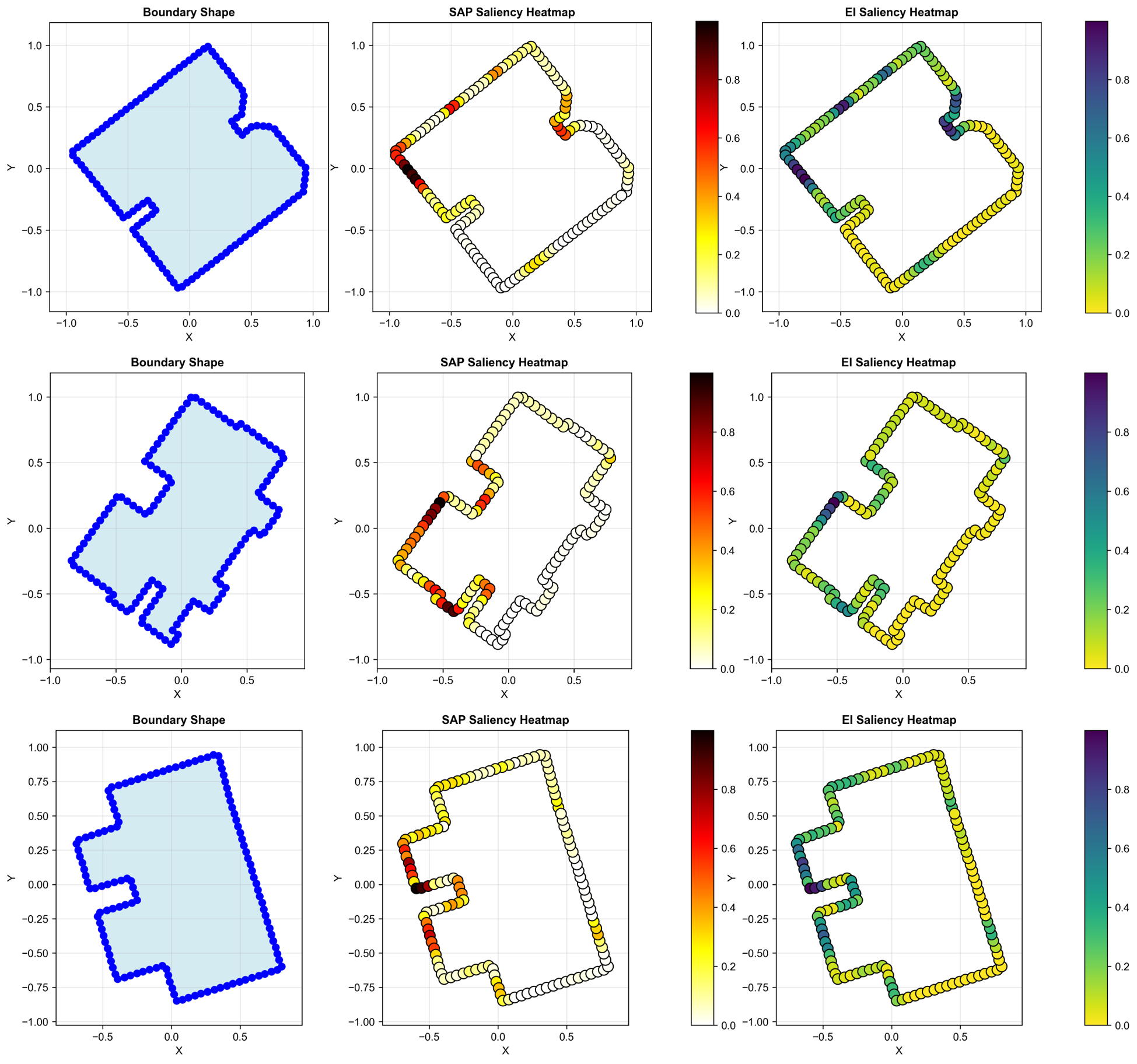}
\caption{Point-level saliency of boundary.}
\label{fig:20}
\end{figure}

Figure~\ref{fig:20} illustrates point-level saliency for three
representative building boundaries. Different boundary points exhibit
varying importance, indicating that the model captures localised
geometric regions that are more sensitive to SAP and EI prediction
rather than treating geometry as a uniformly contributing entity. This
fine-grained geometric interpretability has the potential to support
targeted retrofit considerations by highlighting boundary segments most
influential for energy performance.

\section{Scenario-Based Retrofit Analysis in Westminster}

To demonstrate the practical utility of the proposed framework, three retrofit scenarios were evaluated for properties within Westminster: wall insulation, roof insulation, and window glazing upgrades. Across the study area, 100,701 properties were identified as requiring wall insulation, 22,082 as requiring roof insulation, and 48,788 as requiring glazing upgrades, accounting for 80.5\%, 17.7\%, and 39.0\% of the total property stock, respectively.

\begin{figure}[htbp]
\centering
\includegraphics[width=\linewidth]{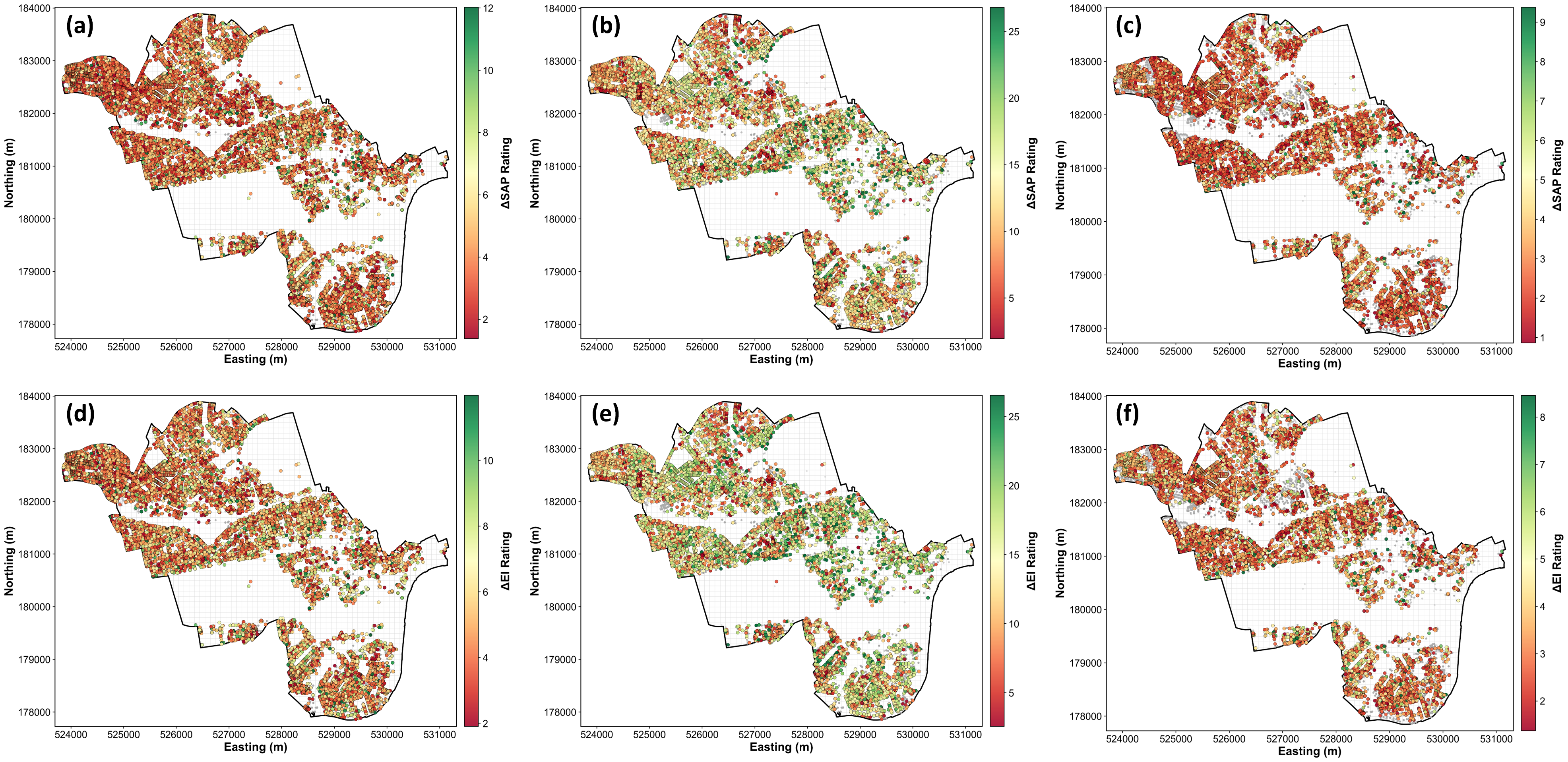}
\caption{Spatial distribution of projected retrofit benefits in Westminster under three intervention scenarios. Panels (a)--(c) show the projected changes in SAP resulting from wall insulation, roof insulation, and window glazing upgrades, respectively, while panels (d)--(f) show the corresponding changes in EI. Properties not requiring the corresponding intervention are shown in grey.}
\label{fig:retrofit_spatial}
\end{figure}

Figure~\ref{fig:retrofit_spatial} illustrates the spatial distribution of properties requiring intervention under each scenario, together with the projected changes in SAP and EI scores after retrofit. Properties not requiring the corresponding intervention are shown in grey. All three retrofit scenarios lead to positive improvements in both SAP and EI. On average, wall insulation increases the SAP and EI scores of the affected properties by 4.64 and 5.66, respectively. The corresponding average gains for roof insulation are 12.01 for SAP and 13.95 for EI, while window glazing upgrades increase SAP and EI by 3.07 and 3.64, respectively. These projected improvements indicate that all three measures can contribute to reducing residential energy demand and associated carbon emissions. The scenario results therefore provide evidence to support Westminster City Council in identifying priority areas for intervention and in setting realistic retrofit targets.

\begin{figure}[htbp]
\centering
\includegraphics[width=\linewidth]{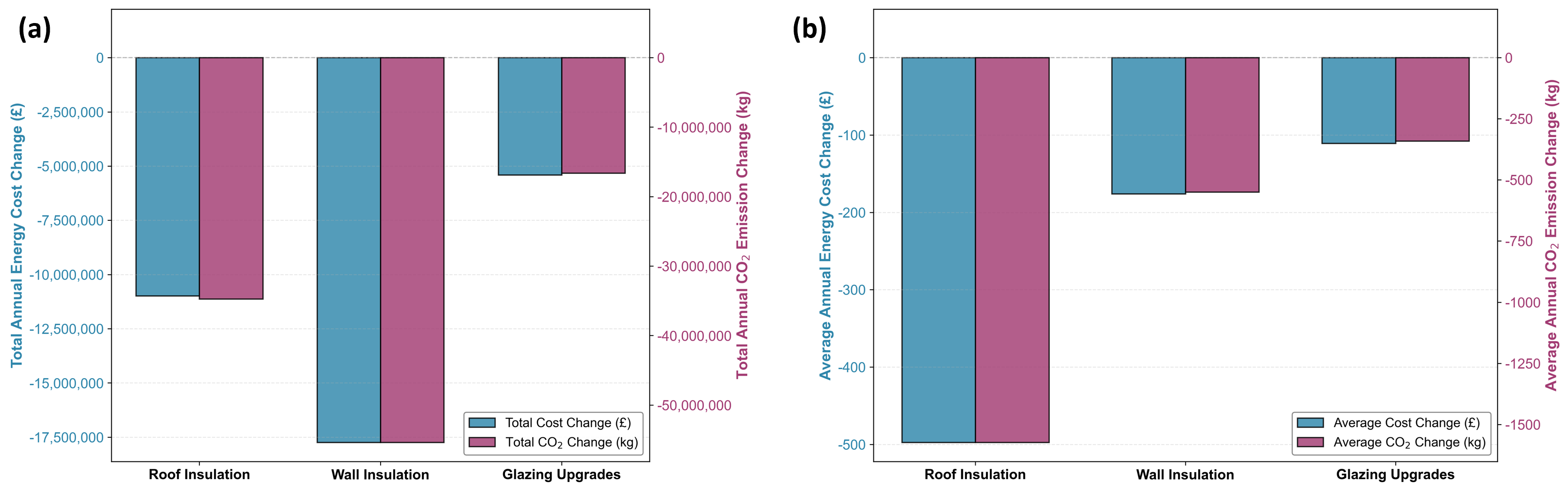}
\caption{Projected annual energy cost and equivalent CO$_2$ emission reductions under three retrofit scenarios in Westminster. Panel (a) shows the total annual reductions aggregated across all properties requiring each intervention, and panel (b) shows the average annual reductions per retrofitted property.}
\label{fig:retrofit_cost_emission}
\end{figure}

Using the score-to-cost and score-to-emission relationships introduced in Section~3.5, the projected SAP and EI improvements were further converted into total annual cost reduction and equivalent CO$_2$ emission reduction, both at the aggregate level and on a per-property basis. The results are shown in Figure~\ref{fig:retrofit_cost_emission}. Retrofitting the 100,701 properties requiring wall insulation is estimated to reduce total annual household energy costs by \pounds17,738,037 and total emissions by 55,367,703~kg~eCO$_2$. For roof insulation, the corresponding total reductions are \pounds10,978,682 and 34,730,080~kg~eCO$_2$ across 22,082 properties. For window glazing upgrades, the projected total reductions are \pounds5,409,651 and 16,617,619~kg~eCO$_2$ across 48,788 properties.

When averaged across the properties receiving each intervention, roof insulation delivers the greatest benefit, with an average annual reduction of \pounds497.18 and 1,572.78~kg~eCO$_2$ per property. This is followed by wall insulation, with average reductions of \pounds176.15 and 549.82~kg~eCO$_2$ per property, and then window glazing upgrades, with corresponding reductions of \pounds110.88 and 340.61~kg~eCO$_2$. These results provide useful evidence for intervention prioritisation. In cases where a property may require multiple retrofit measures, the projected per-property benefits can help inform a more effective sequencing strategy and support the design of retrofit pathways tailored to individual properties.

\section{Conclusion and Future Work}

This study presents an interpretable multimodal framework for large-scale residential energy performance prediction by jointly integrating three complementary data sources: structured EPC tabular features, multi-field EPC text descriptions, and GIS-based spatial information. The proposed model performs dual-target regression for continuous SAP and EI prediction, while an auxiliary band-based classification task is introduced to strengthen supervision. A sample-wise gated fusion mechanism further enables adaptive weighting of modality contributions, improving both predictive accuracy and interpretability. Taken together, the framework provides a scalable pathway for rapid energy performance assessment and retrofit-oriented scenario screening at the urban scale.

The Westminster case study demonstrates that the proposed framework achieves robust and consistent predictive performance for both SAP and EI. The full multimodal configuration consistently outperforms unimodal and bimodal variants in both continuous regression and band-based evaluation, confirming the complementarity of tabular, textual, and spatial information. Subgroup analyses across property types, built forms, and construction age bands further show that the model maintains stable performance across diverse housing categories, with no evidence of systematic degradation. Where score-level performance decreases for specific subgroups, band-level accuracy remains comparatively strong, highlighting the practical value of continuous prediction for capturing fine-grained variations relevant to retrofit planning.

A multi-level interpretability analysis helps explain why the framework performs effectively. At the modality level, the gated fusion weights show that textual information provides the strongest and most consistent contribution, spatial information plays a more heterogeneous and sample-dependent role, and tabular attributes offer a stable supporting baseline. At the feature level, SHAP analysis identifies main fuel, built form, and construction age band as the most influential tabular variables, while field-level text occlusion highlights the importance of fabric-related descriptions, especially roofs and walls. For the spatial modality, permutation analysis shows that building height and footprint area are more informative than orientation, while boundary-based analysis confirms that building geometry contributes complementary information beyond structured and textual inputs.

Beyond prediction, the proposed framework also supports scenario-based retrofit assessment. In Westminster, three retrofit scenarios, including wall insulation, roof insulation, and window glazing upgrades, were evaluated using model-estimated changes in SAP and EI and their corresponding annual cost and emissions implications. All three measures produce positive projected improvements in both indicators. At the aggregate level, wall insulation delivers the largest total reductions because it applies to the widest share of the housing stock, while roof insulation provides the greatest average annual cost and emissions reduction per retrofitted property. These results demonstrate that the framework can support local authorities not only in identifying priority areas for intervention, but also in comparing retrofit options and informing the sequencing of measures at the property level.

Overall, the proposed multimodal framework achieves a strong balance between accuracy, robustness, interpretability, and practical relevance. It enables fine-grained characterisation of residential energy performance and provides a scalable technical basis for rapid retrofit screening, target setting, and evidence-based urban energy governance.

Several limitations should be acknowledged. First, the case study is confined to Westminster, and the transferability of the framework to suburban, rural, or socioeconomically different contexts remains to be tested. Cross-borough and cross-city validation is therefore an important direction for future work. Second, although the multimodal model outperforms reduced-modality variants, future evaluation should include stronger non-neural tabular baselines, such as XGBoost or LightGBM, to more clearly quantify the incremental value of multimodal fusion. Third, the contribution of the boundary-sequence encoder, while consistent, remains modest, and its computational cost-effectiveness warrants further investigation. Fourth, the retrofit analysis presented here is a model-based scenario projection rather than a physical building simulation or a causal estimate of intervention effects; future work should therefore explore tighter integration with engineering simulation and uncertainty analysis. Finally, as the UK assessment framework evolves beyond SAP toward successor methodologies such as the Home Energy Model, adapting the proposed approach to the new policy context will be essential to maintain long-term practical relevance.

\section*{Acknowledgement}

The authors would like to thank the funding support from Westminster
City Council and King's College London.

\bibliographystyle{elsarticle-num}
\bibliography{references}

\end{document}